\documentclass{article} %

\usepackage[accepted]{common/icml2023}

\usepackage{hyperref}
\usepackage{url}

\usepackage[utf8]{inputenc}
\usepackage[T1]{fontenc}
\usepackage{booktabs}
\usepackage{amsfonts}       %
\usepackage{nicefrac}       %
\usepackage{microtype}      %
\usepackage{xcolor}
\usepackage[flushleft]{threeparttable}
\usepackage{color}
\usepackage{mathtools}
\usepackage{amsmath}

\usepackage{url}
\usepackage{float}
\usepackage{subfigure}
\usepackage{graphicx}
\usepackage{multirow}
\usepackage{xspace}
\usepackage{natbib}
\usepackage{enumitem}
\usepackage[font=small]{caption}
\usepackage{diagbox}
\usepackage{wrapfig}
\usepackage[toc, page, header]{appendix}
\definecolor{myblue}{rgb}{0,0.45,0.74}
\definecolor{myred}{rgb}{0.85,0.33,0.1}

\usepackage{amsmath,amsthm,amssymb}
\newtheorem{theorem}{Theorem}[section]

\newtheorem{example}[theorem]{Example}

\newtheorem{proposition}[theorem]{Proposition}

\renewcommand{\gg}{\mathbf{g}}

\renewcommand{\ll}{\mathbf{l}}

\providecommand{\xx}{\mathbf{x}}
\providecommand{\yy}{\mathbf{y}}

\providecommand{\mphi}{\boldsymbol{\phi}}
\providecommand{\momega}{\boldsymbol{\omega}}

\providecommand{\mtheta}{\boldsymbol{\theta}}

\providecommand{\cL}{\mathcal{L}}

\newenvironment{talign*}
{\csname align*\endcsname}
{\endalign}
\usepackage{amssymb}
\usepackage{algorithm}
\usepackage[noend]{algpseudocode}

\newcommand{\mycaptionof}[2]{\captionof{#1}{#2}}

\errorcontextlines\maxdimen

\makeatletter
\newcommand*{\algrule}[1][\algorithmicindent]{\makebox[#1][l]{\hspace*{.5em}\thealgruleextra\vrule height \thealgruleheight depth \thealgruledepth}}%
\newcommand*{\thealgruleextra}{}
\newcommand*{\thealgruleheight}{.75\baselineskip}
\newcommand*{\thealgruledepth}{.25\baselineskip}

\newcount\ALG@printindent@tempcnta
\def\ALG@printindent{%
	\ifnum \theALG@nested>0%
	\ifx\ALG@text\ALG@x@notext%
	\else
		\unskip
		\addvspace{-1pt}%
		\ALG@printindent@tempcnta=1
		\loop
		\algrule[\csname ALG@ind@\the\ALG@printindent@tempcnta\endcsname]%
		\advance \ALG@printindent@tempcnta 1
		\ifnum \ALG@printindent@tempcnta<\numexpr\theALG@nested+1\relax%
			\repeat
		\fi
	\fi
}%
\usepackage{etoolbox}
\patchcmd{\ALG@doentity}{\noindent\hskip\ALG@tlm}{\ALG@printindent}{}{\errmessage{failed to patch}}
\makeatother

\newbox\statebox
\newcommand{\myState}[1]{%
	\setbox\statebox=\vbox{#1}%
	\edef\thealgruleheight{\dimexpr \the\ht\statebox+1pt\relax}%
	\edef\thealgruledepth{\dimexpr \the\dp\statebox+1pt\relax}%
	\ifdim\thealgruleheight<.75\baselineskip
		\def\thealgruleheight{\dimexpr .75\baselineskip+1pt\relax}%
	\fi
	\ifdim\thealgruledepth<.25\baselineskip
		\def\thealgruledepth{\dimexpr .25\baselineskip+1pt\relax}%
	\fi
	\State #1%
	\def\thealgruleheight{\dimexpr .75\baselineskip+1pt\relax}%
	\def\thealgruledepth{\dimexpr .25\baselineskip+1pt\relax}%
}

\newcommand{\algopt}{FedBR\xspace} 

\usepackage[textwidth=5cm]{todonotes}

\icmltitlerunning{\algopt: Improving Federated Learning on Heterogeneous Data via Local Learning Bias Reduction}

\begin{document}

% \twocolumn[\icmltitle{\algopt: Reducing the Local Learning Bias Improves \\ Federated Learning on Heterogeneous Data}

\twocolumn[\icmltitle{\algopt: Improving Federated Learning on Heterogeneous Data \\ via Local Learning Bias Reduction}
    % \algopt: Improving Federated Learning on Heterogeneous Data via Local Learning Bias Reduction
    % \algopt: Improving Federated Deep Learning on Heterogeneous Data via Local Learning Bias Reduction
    % ?

    \icmlsetsymbol{equal}{*}

    \begin{icmlauthorlist}
        \icmlauthor{Yongxin Guo}{CUHKSZ,AIRS}
        \icmlauthor{Xiaoying Tang}{CUHKSZ,AIRS,FNI}
        \icmlauthor{Tao Lin}{WestlakeRCIF,WestlakeSoE}
    \end{icmlauthorlist}

    \icmlaffiliation{CUHKSZ}{School of Science and Engineering, The Chinese University of Hong Kong, Shenzhen, Guangdong 518172, China.}
    \icmlaffiliation{WestlakeRCIF}{Research Center for Industries of the Future, Westlake University}
    \icmlaffiliation{WestlakeSoE}{School of Engineering, Westlake University}
    \icmlaffiliation{AIRS}{The Shenzhen Institute of Artificial Intelligence and Robotics for Society}
    \icmlaffiliation{FNI}{The Guangdong Provincial Key Laboratory of Future Networks of Intelligence, The Chinese University of Hong Kong (Shenzhen), Shenzhen 518172, China }

    \icmlcorrespondingauthor{Xiaoying Tang}{tangxiaoying@cuhk.edu.cn}

    % You may provide any keywords that you
    % find helpful for describing your paper; these are used to populate
    % the "keywords" metadata in the PDF but will not be shown in the document
    \icmlkeywords{Machine Learning, ICML}
    \vskip 0.3in]

\printAffiliationsAndNotice{}
% \printAffiliationsAndNotice{\icmlEqualContribution} % otherwise use the standard text.

%%%%%%%%%%%%%%%%%%%%%%%%%%%%%%%%%%%%%%%%%%%%%%%%%%%%%%%%%%%%
% !TeX root = icml2023_fl_invariant.tex
% \setlength{\parskip}{1.75pt plus1.75pt minus0pt}
\setlength{\parskip}{0pt plus0pt minus0pt}

\begin{abstract}
	Federated Learning (FL) is a way for machines to learn from data that is kept locally, in order to protect the privacy of clients. This is typically done using local SGD, which helps to improve communication efficiency. However, such a scheme is currently constrained by slow and unstable convergence due to the variety of data on different clients' devices. In this work, we identify three under-explored phenomena of biased local learning that may explain these challenges caused by local updates in supervised FL.
	As a remedy, we propose \algopt, a novel unified algorithm that reduces the local learning bias on features and classifiers to tackle these challenges.
	\algopt has two components. The first component helps to reduce bias in local classifiers by balancing the output of the models. The second component helps to learn local features that are similar to global features, but different from those learned from other data sources. We conducted several experiments to test \algopt and found that it consistently outperforms other SOTA FL methods. Both of its components also individually show performance gains.
	Our code is available at \url{https://github.com/lins-lab/fedbr}.

	% Federated Learning (FL) is a machine learning paradigm that learns from data kept locally to safeguard the privacy of clients, whereas local SGD is typically employed on the clients' devices to improve communication efficiency. However, such a scheme is currently constrained by the slow and unstable convergence induced by clients' heterogeneous data.
	% In this work, we identify three under-explored phenomena of biased local learning that may explain these challenges caused by local updates in supervised FL.
	% As a remedy, we propose \algopt, a novel unified algorithm that reduces the local learning bias on features and classifiers to tackle these challenges.
	% \algopt consists of two components:
	% The first component alleviates the bias in the local classifiers by balancing the output distribution of models.
	% The second component learns local features that are close to global features but considerably distinct from those learned from other input distributions.
	% In a series of experiments, we show that \algopt consistently outperforms other SOTA FL baselines, in which both two components have individual performance gains.
	\looseness=-1

\end{abstract}

\section{Introduction}
% \paragraph{Federated Learning and the challenge of heterogeneous data.} 
Federated Learning (FL) is an emerging privacy-preserving distributed machine learning paradigm.
The model is transmitted to the clients by the server, and when the clients have completed local training, the parameter updates are sent back to the server for integration.
Clients are not required to provide local raw data during this procedure, maintaining their privacy.
As the workhorse algorithm in FL, FedAvg \citep{mcmahan2016communication} proposes local SGD to improve communication efficiency.
However, the considerable heterogeneity between local client datasets leads to inconsistent local updates and hinders convergence.

Several studies propose variance reduction methods \citep{karimireddyscaffold2019,das2020faster}, or suggest regularizing local updates towards global models \citep{lifederated2018,li2021model} to tackle this issue.
Almost all these existing works directly regularize models by utilizing the global model collected from previous rounds to reduce the variance or minimize the distance between global and local models~\citep{lifederated2018,li2021model}.
However, it is hard to balance the trade-offs between optimization and regularization to perform well, and data heterogeneity remains an open question in the community, as justified by the limited performance gain, e.g.\ in our Table~\ref{Performance of algorithms} and experiment results in some previous works~\citep{tang2022virtual,li2021model,yoon2021federated,chen2021bridging,luo2021no}.
\looseness=-1
% due to the diminished regularization objective during training.
% \tao{support for the claim of ``diminished regularization objective during training''}.
% % However, such a design, by regularizing models toward the global model collected from previous rounds---which is also the initial model of the current round---will hinder the convergence of the current round. However, it is hard to balance the trade-offs between optimization and regularization to perform well.
% However, the performance of these algorithms are not strong as theoretically proof in practice, \tang{risky..Is there any analysis or experiments to support this statement? Any references?}
% Therefore, we are ready to develop an efficient algorithm to further enhance the performance. 

% Yet there exist no mechanism to balance the trade-offs between optimization and regularization.
% For example, when the weights of proximal terms become too large in FedProx~\cite{lifederated2018}, local models could not converge.
% \tao{?} \guo{add an example}
% Instead, we treat the global features as a class of features, and local features as another class, then 
% However, most existing works only consider minimizing the distance between global and local models, and use global model of previous rounds to represent the global model in current round.\tao{again this statement..} 

\begin{figure*}
	\centering
	\subfigure[\small Bias in local classifier.]{\includegraphics[width=0.475\textwidth]{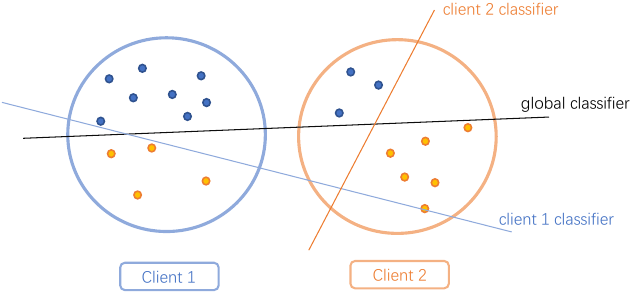}
		\label{fig:draft1}
	}
	\hfill
	\subfigure[\small Bias in local feature.]{\includegraphics[width=0.4\textwidth]{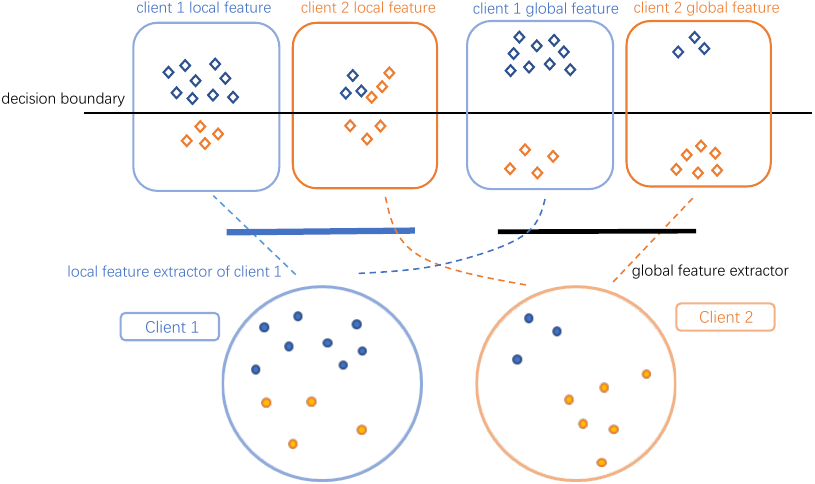}
		\label{fig:draft2}
	}
	\vspace{-1.25em}
	\caption{\small
		\textbf{Observation for learning bias of FL} on heterogeneous data with local updates.
		There are two clients in the figure, and each has two classes of data (red and blue points).
		% The bias in local classifiers is shown in subfigure
		\textbf{Biased local classifier} in~\ref{fig:draft1}: Client 1's decision boundary cannot accurately classify data samples from Client 2.
		% Local feature bias is depicted in subfigure (b). 
		\textbf{Biased local feature} in~\ref{fig:draft2}:
		The difference between features extracted by Client 1's local and global feature extractor is sustainable large.
		However, Client 2's local feature is close enough to Client 1's, even for input data from different data distributions/clients.
		\looseness=-1
	}
	\label{learning bias in FL}
	% \vspace{-1.75em}
\end{figure*}

Apart from the existing solutions, we revisit and reinterpret the fundamental issues in federated deep learning, caused by data heterogeneity and local updates.
As our first contribution, we identify three pitfalls of FL systematically and in a unified view, termed \emph{local learning bias}, from the perspective of representation learning\footnote{Please refer to section \ref{sec:The Pitfalls of FL on Heterogeneous Data Distributions} for more justifications about the existence of our observations.}:
1) Biased local classifiers are unable to effectively classify unseen data (c.f.\ Figure~\ref{fig:draft1}), due to the shifted decision boundaries dominated by local class distributions; 2) Local features (extracted by a local model) differ significantly from global features (similarly extracted by a centralized global model), even for the same input data (c.f.\ Figure~\ref{fig:draft2}); and 3) Local features, even for data from different classes, are too close to each other to be accurately distinguished (c.f.\ Figure~\ref{fig:draft2}).
\looseness=-1

% \begin{wrapfigure}{R}{0.35\textwidth}
% \guo{not sure if this figure should be saved, for it contain the same information as figure 3, 4.}
% \end{wrapfigure}

% \paragraph{\algopt as a crucial step to remedy the learning bias in FL.}
To this end, we propose \algopt, a unified method that leverages (1) a globally shared, label-distribution-agnostic pseudo-data and (2) two key algorithmic components, to simultaneously address the three difficulties outlined above. \\
The first component of \algopt alleviates the first difficulty by forcing the output distribution of the pseudo-data to be balanced.
The second component of \algopt aims to address the second and third difficulties simultaneously, where we develop a \emph{\textbf{min-max contrastive learning method}} to learn client invariant local features.
More precisely, a key two-stage algorithm is proposed to maximize the elimination of feature learning biases: the first stage learns a projection space to distinguish the features of two types, while the second stage enforces learned features on the projected feature space that are farther from local features and closer to global ones.
All these can be unified into a simple yet effective min-max procedure to alleviate the local learning bias in FL, with trivial requirements on pseudo-data while still preserving privacy.
\looseness=-1

% We examine the performance of \algopt and compare it with other FL baselines on RotatedMNIST, CIFAR10, and CIFAR100.
% The numerical results indicate that \algopt significantly outperforms other algorithms in terms of both mean accuracy and convergence speed.
% Additionally, we show the well-designed min-max process is necessary for \algopt to achieve good performance.
% \looseness=-1

\paragraph*{Our main contributions are:}
\begin{itemize}[leftmargin=12pt,nosep]
	\item We provide a unified view to interpret the learning difficulty in FL with heterogeneous data, and identify three key pitfalls to explain the issue of local learning biases.
	\item We propose \algopt, a unified algorithm that leverages pseudo-data to reduce the learning bias on local features and classifiers.
	      We design two orthogonal key components of \algopt to complement each other to improve the learning quality of clients with heterogeneous data.
	      % \item AugMean is a simple but effective method for balancing the local output distribution of augmentation pseudo-data in order to eliminate local classifier bias;
	      % \item AugCA first use a projection operation to distinguish between global and local features. Then learn features closer to global features, and far from local features on the projection spaces to reduce the bias in local features. The results suggest that adopting projection space improves algorithm performance significantly.
	\item \algopt considerably outperforms other FL baselines by a large margin, as justified by extensive numerical evaluation on RotatedMNIST, CIFAR10, and CIFAR100. Besides, \algopt does not require the labeled or large number of global shared pseudo-data, thereby improving the efficiency.
\end{itemize}
% \begin{itemize}[leftmargin=12pt,nosep]
% 	\item We identify three challenges that may explain the learning difficulty cased by local updates: 1) the biased local classifiers; 2) difference between local and global features for same inputs; 3) similarity of local features for different inputs.\tao{it is a bit strange to put them as a contribution...but i also don't know how to improve...} 
% 	To the best of our knowledge, we are the first to examine the third obstacle and provide a solution to address all three challenges simultaneously. \tao{also duplicated with the second point?}
% 	\item We propose \algopt to address the three difficulties, which can be divided into two components: AugMean and AugCA. 
% 	AugMean solves the first difficulty, and AugCA overcomes the others by constructing an contrastive adversarial problem to learn client invariant features. 
% 	\item We evaluate \algopt on MNIST, CIFAR10, and PACS. 
% 	Results show \algopt outperform other SOTA FL and domain generalization methods significantly, and both AugMean and AugCA have improvements compare with FedAvg. 
% 	In addition, \algopt outperform other baselines both on mean accuracy and worst case accuracy, which means the method can also improve the distribution robustness.
% \end{itemize}

\section{Related Works}

\paragraph{Federated Learning (FL).}
As the de facto FL algorithm, \citet{mcmahan2016communication,lin2020dont} propose to use local SGD steps to alleviate the communication bottleneck.
However, the objective inconsistency caused by the local data heterogeneity considerably hinders the convergence of FL algorithms~\citep{lifederated2018,wang2020tackling,karimireddyscaffold2019,karimireddy2020mime,guo2021towards}.
% Studies show that the average of local models could be far away from the true global models that centralized trained~\cite{zhao2018federated}, and the difference increase when the number of local updates becomes large~\cite{karimireddyscaffold2019,sahu2018convergence,wang2020federated}. 
To address the issue of heterogeneity in FL, a series of projects has been proposed. FedProx~\citep{lifederated2018} incorporates a proximal term into local objective functions to reduce the gap between the local and global models. SCAFFOLD~\citep{karimireddyscaffold2019} adopts the variance reduction method on local updates, and Mime~\citep{karimireddy2020mime} increases convergence speed by adding global momentum to global updates. Recently, Moon~\citep{li2021model} has proposed to employ contrastive loss to reduce the distance between global and local features.
However, their projection layer is only used as part of the feature extractor, and cannot contribute to distinguishing the local and global features---a crucial step identified by our investigation for better model performance.\looseness=-1

In this paper, we focus on improving the global model in Federated Learning (FL) by designing methods that perform well on all local distributions. The designed algorithm works on the local training stage, which aligns with previous research in this area, such as~\citet{mcmahan2016communication, lifederated2018, karimireddyscaffold2019, li2021model, tang2022virtual}.
Other topics like improving the global aggregation stages~\citep{wang2020federated, yoshida2019hybrid} or Personalized Federated Learning (PFL) methods~\citep{tan2022towards, wu2022motley,jiang2023testtime} are orthogonal to our approach and could be further combined with our method.
% Our research does not focus on other studies that improve global aggregation stages~\citep{wang2020federated, yoshida2019hybrid} or Personalized Federated Learning (PFL) methods which aim to learn distinct local models for each client~\citep{tan2022towards, wu2022motley}. However, our method could potentially be combined with PFL methods or more advanced global aggregation approaches.

% Other studies that improve global aggregation stages~\citep{wang2020federated, yoshida2019hybrid} or Personalized Federated Learning (PFL) methods, which aim to learn distinct local models for each client~\citep{tan2022towards, wu2022motley} are not the focus of our research.
% However, our method can potentially be combined with PFL methods~\citep{wu2022motley} or more advanced global aggregation approaches~\citep{wang2020federated, yoshida2019hybrid}.\tao{try to polish and shorten this part (due to the duplication).}

\paragraph{Data Augmentation in FL.}
To reduce data heterogeneity, some data-based approaches suggest sharing a global dataset among clients and combining global datasets with local datasets~\citep{tuor2021overcoming,yoshida2019hybrid}.
Some knowledge distillation-based methods also require a global dataset~\citep{tao2020ensemble,li2019fedmd}, which is used to transfer knowledge from local models (teachers) to global models (students).
Considering the impractical of sharing the global datasets in FL settings, some recent research use proxy datasets with augmentation techniques.
Astraea~\citep{duan2019astraea} uses local augmentation to create a globally balanced distribution.
XorMixFL~\citep{shin2020xor} encodes a couple of local data and decodes it on the server using the XOR operator.
FedMix~\citep{yoon2021fedmix} creates the privacy-protected augmentation data by averaging local batches and then applying Mixup in local iterations.
% In this work, we use the augmentation framework in FedMix as an option to construct the pseudo-data in our framework, and use the pseudo-data to reduce the local learning bias. 
% In the experiment section, we show the significant performance gain of our method compare with FedMix using the same augmentation framework.
VHL~\citep{tang2022virtual} relies on the created virtual data with labels and forces the local features to be close to the features of same-class virtual data.
Different from previous works, this paper designs methods that utilize label-agnostic pseudo-data, and outperform other methods using significantly less pseudo-data.

\begin{figure*}[!t]
	\centering
	\subfigure[\small Global feature of $X_1$, $F_g(X_1)$]{\includegraphics[width=0.32\textwidth]{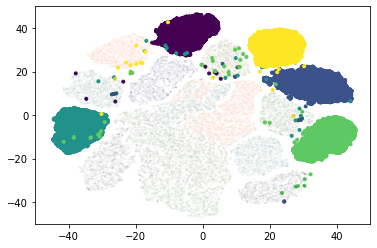}
		\label{global_feature}
	}
	\subfigure[\small Local feature of $X_1$, $F_1(X_1)$]{\includegraphics[width=0.32\textwidth]{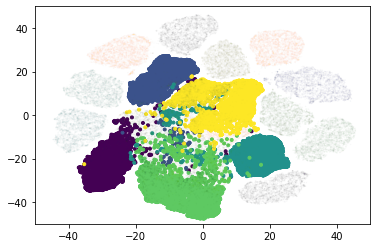}
		\label{local_feature_1}
	}
	\subfigure[\small Local feature of $X_2$, $F_1(X_2)$]{\includegraphics[width=0.32\textwidth]{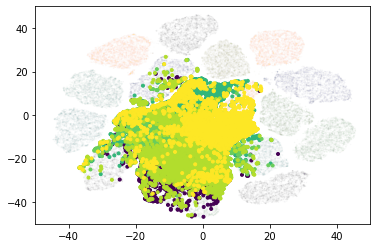}
		\label{local_feature_2}
	}
	\vspace{-1.25em}
	\caption{\small
		\textbf{Observation for biased local features} on a shared t-SNE projection space.
		\emph{Local updates will cause: $\bullet$ Large difference in local and global features for the same input data}.
		Colored points in sub-figures (a) \& (b) denote the global and local features of data from $X_1$, and the same color indicates data from the same class.
		Notice that even for data from the same class (same color), the global and local features are clustered into two distinct groups, implying a considerable distance between global and local features even for the same input data distribution.
		\emph{$\bullet$ High similarity of local features for different inputs.}
		Notice from sub-figure (b) \& (c) that $X_1$ and $X_2$ are two disjoint datasets (no data from the same class).
		However, the local features of $X_1$ and $X_2$ are clustered into the same group by t-SNE, indicating the relatively small distance between local features of different classes. Results on mild conditions and different training stages can be found in Appendix~\ref{sec:Additional-Results}.
		% 		We use t-SNE~\citep{van2008visualizing}, a unsupervised cluster algorithm, to project features onto 2D spaces. 
		% 		To promise the fairness, we project $F_g(X_1)$, $F_g(X_2)$, $F_1(X_1)$, and $F_1(X_2)$ onto same spaces, where $X_1$ and $X_2$ are two disjoint parts of MNIST contain different classes. $F_1$ is the local feature extractor trained on data $X_1$, and $F_g$ is the global feature extractor centralized trained on the whole MNIST dataset.
		% 		We draw points for each feature, using solid points to highlight those we want to exhibit and dash points for the others.
		% 		Sub-figures illustrate $F_g(X_1)$, $F_1(X_1)$, and $F_1(X_2)$ respectively.
		%         We found in sub-figure (a) and (b) that the \emph{global and local features are clustered into different groups even for same inputs}.
		%         In sub-figure (c), we notice that even though the points in $F_1(X_2)$ are close to points in $F_1(X_1)$, the data in $X_1$ and $X_2$ indeed belongs to different classes, and should have dramatic difference on features in the expectation.
		\looseness=-1
	}
	% \vspace{-1.25em}
	\label{Output of model trained on class 1-5}
\end{figure*}

\section{The Pitfalls of FL on Heterogeneous Data}
\label{sec:The Pitfalls of FL on Heterogeneous Data Distributions}
\paragraph{FL and local SGD.}
FL is an emerging learning paradigm that supposes learning on various clients while clients can not exchange data to protect users' privacy.
Learning occurs locally on the clients, while the server collects and aggregates gradient updates from the clients.
The standard FL considers the following problem:
\begin{align}
	\textstyle
	f^* = \min_{\momega \in R^d} \left[ f(\momega) = \sum_{i=1}^{N} p_i f_i(\momega) \right] \,,
\end{align}
where $f_i(\momega)$ is the local objective function of client $i$, and $p_i$ is the weight for $f_i(\momega)$. In practice, we set $p_i = \nicefrac{|D_i|}{|D|}$ by default, where $D_i$ is the local dataset of client $i$ and $D$ is the combination of all local datasets. The global objective function $f(\momega)$ aims to find $\momega$ that can perform well on all clients.
\looseness=-1

In the training process of FL, the communication cost between client and server has become an essential factor affecting the training efficiency.
Therefore, local SGD \citep{mcmahan2016communication} has been proposed to reduce the communication round. In local SGD, clients perform multiple local steps before synchronizing to the server in each communication round.
\looseness=-1
% \tao{introduce what is federated learning and the importance of using local SGD in Federated Learning.}

% \tao{use a table/paragraph to argue that most existing distributionally robust methods cannot be applied to federated learning directly (the details can be deferred to appendix, but the key messages should be properly delivered in the main text; some numerical results should be provided or argued here to support the argument).}

% \paragraph{Challenges in a nutshell.}
% In our observation, there are two\tao{it becomes 2 instead 3 mentioned in the abs/intro.} kinds of difficulties that caused by local updates:
% \begin{itemize}[nosep, leftmargin=12pt]
% 	\item Biased local features. Local updates will cause: 1) difference of local and global features for same inputs; 2) similarity of local features for different inputs.
% 	\item Biased local classifiers. The output distribution of the local classifiers will be affected by the local class distribution.
% \end{itemize}

\paragraph{Bias caused by local updates.}
In this paper, we consider improving previous works by proposing a label-agnostic method, and we first identify the pitfalls of FL on heterogeneous data in a label-agnostic way as follows.
% To mitigate the negative impact of local updates, we first identify the pitfalls of FL on heterogeneous data with a sufficient number of local updates and then design the algorithms to address the issues caused by the local updates.
\looseness=-1
% Given that the global model work well on the whole dataset, and does not have the issue of local updates, we consider the $F_g$ as the ground truth model. 

\begin{proposition}[Local Learning Bias in FedAvg] For FedAvg, the local models after local epochs could be biased, in detail,
	\begin{itemize}[leftmargin=12pt,nosep]
		\item \emph{Biased local feature:} For local feature extractor $F_i(\cdot)$, and centralized trained global feature extractor $F_g(\cdot)$, we have:
		      % 1) for input $X$, the difference between $F_i(X)$ and $F_g(X)$ become large. 
		      % 2) For input from different data distribution $X_1$ and $X_2$, the difference between $F_i(X_1)$ and $F_i(X_2)$ could be small. 
		      % \item After doing a sufficient number of local updates, local models classify all samples into the classes only shown in local datasets.
		      1) Given the data input $X$, $F_i(X)$ could deviate largely from $F_g(X)$.
		      2) Given the input from different data distributions $X_1$ and $X_2$, $F_i(X_1)$ could be very similar or almost identical to $F_i(X_2)$.
		\item \emph{Biased local classifier:} After a sufficient number of iterations, local models classify all samples into only the classes that appeared in the local datasets.
	\end{itemize}
	\label{bias caused by local updates}
\end{proposition}

To verify the correctness of Proposition~\ref{bias caused by local updates}, we can use some toy examples to show the existence of the biased local feature and classifiers. For toy examples on more complex scenarios and on the benefits of using \algopt please refer to Appendix~\ref{sec:t-SNE and classcifier output}.

\begin{figure}[!t]
	% \vspace{-1em}
	\centering
	\subfigure[Sample from class 8]{\includegraphics[width=0.23\textwidth]{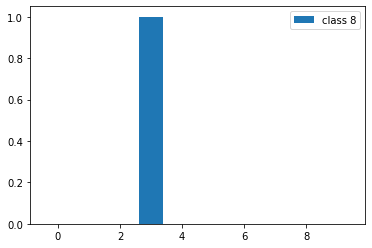}
		\label{output-class-8}
	}
	\subfigure[All samples in $X_2$]{\includegraphics[width=0.23\textwidth]{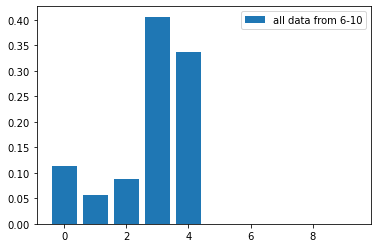}
		\label{output-all-data}
	}
	\vspace{-1.25em}
	\caption{\small
		\textbf{Observation for biased local classifiers}: \emph{the output distribution of the local classifiers will be dominated by the local class distribution}.
		The model is trained on data $X_1$ and tested on data $X_2$. The sub-figure (a) illustrates the model output distribution of a sample belonging to Class 8. The sub-figure (b) shows the total prediction distribution of all samples in $X_2$.
		Results show that \emph{the biased local model will classify all samples into classes that are only present in the $X_1$.}
		\looseness=-1
	}
	% \vspace{-0.5em}
	\label{Class imbalance figure}
\end{figure}

\begin{example} [Observation for biased local features]
	Figures~\ref{global_feature} and~\ref{local_feature_1} show that \emph{local features differ from global features for the same input}, and Figures~\ref{local_feature_1} and~\ref{local_feature_2} show that \emph{local features are similar even for different input distributions.}
	We define this observation as the ``biased local feature''.
	% in Proposition~\ref{bias caused by local updates}.
	% Figure~\ref{Output of model trained on class 1-5} corresponds to the global features of $X_1$, and local features of $X_1$ and $X_2$. 
	In detail, we calculate $F_1(X_1)$, $F_1(X_2)$, $F_g(X_1)$, and $F_g(X_2)$, and use t-SNE to project all the features to the same 2D space.
	% \footnote{
	% 	We provide the results after using \algopt in Appendix~\ref{sec:t-SNE and classifier output}.
	% }
	We can observe that the local features of data in $X_2$ are so close to local features of data in $X_1$, and it is non-trivial to tell which category the current input belongs to by merely looking at the local features.
	% Since the features of data in $X_2$ are so close to those of data in $X_1$, it is non-trivial to tell which category the current input belongs to just by merely looking at the local features. \footnote{We provide the results after using \algopt in Appendix~\ref{sec:t-SNE and classcifier output}.}
	\label{example 1}
\end{example}
\begin{example} [Observation for biased local classifiers]
	Figure \ref{Class imbalance figure} shows the output of the local model on data $X_2$, where all data in $X_2$ are incorrectly categorized into classes 0 to 4 of $X_1$.
	The observation, i.e.\ data from classes that are absent from local datasets cannot be correctly classified by the local classifiers, refers to the ``biased local classifiers''.
	More precisely, Figure~\ref{output-class-8} shows the prediction result of one sample (class $8$) and Figure~\ref{output-all-data} shows the predicted distribution of all samples in $X_2$.
	% 	The results show that all data in $X_2$ are incorrectly categorized into classes 0 to 4 of $X_1$. 
	% 	Consequently,  data from classes that are absent from local datasets cannot be correctly classified by local classifiers.
	\looseness=-1
	\label{example 2}
\end{example}
%  \tang{suggest delete `local updates ...inputs'', since this paragraph is two long}

% \begin{wrapfigure}{R}{0.6\textwidth}
% \end{wrapfigure}

% Based on Example \ref{example 1} and Example \ref{example 2}, we summarize our observations and introduce the formal definition of ``local learning bias'' caused by local updates:

\paragraph{Distinct from the local learning bias in previous works.}
We acknowledge the discussion of learning bias in some previous works, e.g.\ in~\citet{karimireddyscaffold2019,lifederated2018,li2021model}.
However, our work differs in several ways:
\begin{enumerate}[leftmargin=12pt,nosep]
	\item FedProx~\citep{lifederated2018} defines local drifts as the differences in model weights, while SCAFFOLD~\citep{karimireddyscaffold2019} considers gradient differences as client drifts.
	      These methods, though have been effective on traditional optimization tasks, may only have marginal improvements on deep models, as shown in~\citet{tang2022virtual,li2021model,yoon2021federated,chen2021bridging,luo2021no}. \looseness=-1
	\item MOON~\citep{li2021model} minimizes the distance between global and local features, but its performance is limited because they use only the projection layer as part of the feature extractor, and the contrastive loss diminished without our designed max step (c.f.\ Table~\ref{Performance of algorithms} and Table~\ref{ablation study}).
	\item VHL~\citep{tang2022virtual} defines local learning bias as the shift in features between samples of the same classes; however, this approach requires prior knowledge of local label information and results in a much larger virtual dataset, especially when increasing the number of classes.
	      Our method instead achieves better performance with significantly fewer pseudo-data (see Table~\ref{Comparison with VHL}).
\end{enumerate}

% \begin{definition} [Biasd local classifiers]
%     After doing a sufficient number of local updates, local models classify all samples into the classes shown in local datasets.
%     \label{Personalization of local classification}
% \end{definition}

\begin{figure}[!t]
	\centering
	% \vspace{-0.5em}
	\includegraphics[width=0.35\textwidth]{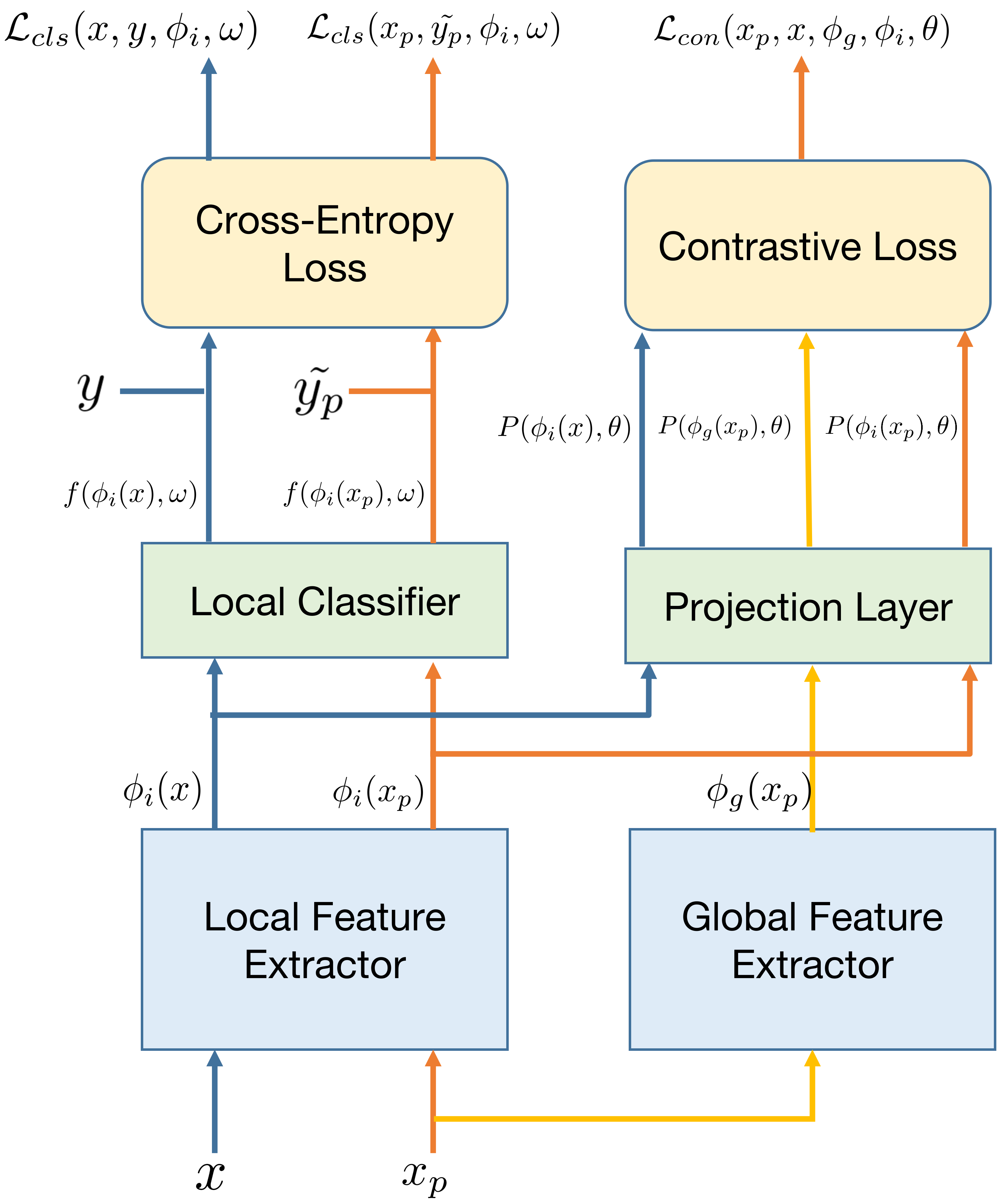}
	\vspace{-1.em}
	\caption{\small
		\textbf{Optimization flow of \algopt}: an illustration of how the three terms in~\eqref{adversary step equation} and \eqref{classification step eqation} are calculated.
		% an illustration of calculating the three terms in~\eqref{adversary step equation} and \eqref{classification step eqation}.
		We calculate the cross-entropy loss of local data $(\xx, \yy)$, and pseudo-data $(\xx_p, \tilde{\yy_p})$, and use the local feature $\mphi_i(\xx), \mphi_i(\xx_p)$, and global feature $\mphi_g(\xx_p)$ for contrastive loss.
		\looseness=-1
	}
	% \vspace{-1.75em}
	\label{Overview of FedAug}
\end{figure}

\section{\algopt: Reducing Learning Bias in FL}
Addressing the local learning bias is crucial to improving FL on heterogeneous data, due to the \emph{bias} discussed in Proposition~\ref{bias caused by local updates}.
To this end, we propose \algopt as shown in Figure~\ref{Overview of FedAug}, a novel framework that leverages the globally shared pseudo-data with two key components to reduce the local training bias, namely
1) reducing the local classifier's bias by balancing the output distribution of classifiers (component 1),
% (The \textcolor[HTML]{ed6d00}{orange} line in Figure~\ref{Overview of FedAug} that obtain $\cL_{cls} (\xx_p, \tilde{\yy_p}, \mphi_i, \momega)$)
and 2) an adversary contrastive scheme to learn unbiased local features (component 2).
% (The right side of Figure~\ref{Overview of FedAug} that obtain $\cL_{con} (\xx_{p}, \xx, \mphi_g, \mphi_i, \boldsymbol{\theta})$). 

% To overcome the ``bias'' in Proposition~\ref{bias caused by local updates}, it is natural to offer strategies for jointly addressing the problem of biased feature and classifier. In this end, we propose a novel framework, named \algopt, which use global shared pseudo-data to reduce the local bias in FL training. 
% In this section, we first show how to build the pseudo-data. Then we design two methods to jointly address the  ``bias'' in Proposition~\ref{bias caused by local updates}: 1) Reduce the bias in local classifier by classify the pseudo-data that does not belong to any classes. 2) An adversary contrastive method to overcome the bias in local feature. 

\subsection{Overview of the \algopt}\label{sec:overview}
The learning procedure of \algopt on each client $i$ involves the construction of a global pseudo-data (c.f.\ Section~\ref{sec:construction}), followed by applying two key debias steps in a \textbf{\emph{min-max}} approach to jointly form two components (c.f.\ Section~\ref{sec:component1} and~\ref{sec:component2}) to reduce the bias in the classifier and feature, respectively.
% We construct an adversary problem that first maximize the contrastive loss, then minimize the classification loss together with the contrastive loss:
% In this section, we investigate how to use the pseudo-data to overcome the learning bias in FL training. In this end, we propose a two-stage method that jointly addressing the problem of biased feature and classifier as illustrated in Proposition~\ref{bias caused by local updates}. The overview of the proposed algorithm on client $i$ is as follows, and the algorithm procedure is the same on all clients:

The min-max procedure of \algopt can be interpreted as first projecting features onto spaces that can distinguish global and local feature best, then 1) minimizing the distance between the global and local features of pseudo-data and maximizing distance between local features of pseudo-data and local data; 2) minimize classification loss of both local data and pseudo-data:
% We construct an adversary problem that first maximize the contrastive loss, then minimize the classification loss together with the contrastive loss:

\textbf{Max Step:} $\max_{\boldsymbol{\theta}} \cL_{adv} (D_p, D_i) $
\begin{small}
	\begin{align}
		% \max_{\boldsymbol{\theta}} \cL_{adv} (D_p, D_i)
		:= \mathbb{E}_{\xx_p \sim D_p, \xx \sim D_i} \left[ \cL_{con} (\xx_{p}, \xx, \mphi_g, \mphi_i, \boldsymbol{\theta}) \right] \,.
		\label{adversary step equation}
	\end{align}
\end{small}
\textbf{Min Step:} $\min_{\mphi_i, \momega} \cL_{gen} (D_p, D_i)$
\begin{small}
	\begin{align}
		 & := \mathbb{E}_{(\xx, \yy) \sim D_i} \left[ \cL_{cls} (\xx, \yy, \mphi_i, \momega) \right] \nonumber                                                                           \\
		 & \qquad + \lambda \mathbb{E}_{\xx_p\sim D_p} \left[ \cL_{cls} (\xx_p, \tilde{\yy_p}, \mphi_i, \momega) \right] \nonumber                                                       \\
		 & \qquad + \mu \mathbb{E}_{\xx_p \sim D_p, \xx \sim D_i} \left[ \cL_{con} (\xx_{p}, \xx, \mphi_g, \mphi_i, \boldsymbol{\theta}) \right] \,. \label{classification step eqation}
	\end{align}
\end{small}

$\cL_{cls}$ and $\cL_{con}$ represent the cross-entropy loss and a contrastive loss (will be detailed in Section~\ref{sec:component2}), respectively.
$D_i$ denotes the distribution of the local dataset at client $i$.
$D_p$ is that of shared pseudo-dataset, where $\tilde{\yy_p}$ is the pseudo-label of pseudo-data.
The model is composed of a feature extractor $\mphi$ and a classifier $\momega$, where the omitted subscript $i$ and $g$ correspond to the local client $i$ and global parameters, respectively (e.g.\ $\mphi_g$ denotes the feature extractors received from the server at the beginning of each communication round).
We additionally use a projection layer $\boldsymbol{\theta}$ for the max step to project features onto spaces where global and local features have the largest dissimilarity.\\
Apart from the cross-entropy loss of local data in~\eqref{classification step eqation}, the second term aims to overcome the biased local classifier while the local feature is debiased by the third term.
\looseness=-1

The proposed \algopt is summarized in Algorithm \ref{Algorithm Framework of AugCA}.
% the algorithm updates projection layer $\boldsymbol{\theta}$ in the max step to maximize $\cL_{adv}$, while in the min step, the algorithm updates model parameters (i.e.\ $\mphi$ and $\momega$) by minimizing $\cL_{gen}$.\tao{this sentence is somehow useless.}
The global communication part is the same as FedAvg,
% We have just included the gradient descent version of the algorithm, and it can be easily extended to the mini-batch SGD version. 
and the choice of synchronizing the new pseudo-data to clients in each round is optional\footnote{
	As shown in Figure~\ref{Performance when only transfer pseudo-data at the beginning of training}, the communication-efficient variant of \algopt---i.e.\ only transferring pseudo-data at the beginning of the FL training---is on par with the choice of frequent pseudo-data synchronization.
}.

\paragraph{The benefit of \algopt for using less prior information.} In additional to the superior performance, the design of \algopt has the following benefits:
1) Unlike previous works~\cite{tang2022virtual}, our method does not require knowledge of the local label distribution and is label-agnostic.
This means that the size of the pseudo-data will not increase as the number of classes increases.
Our results, shown in Table~\ref{Comparison with VHL} and Figure~\ref{Performance when only transfer pseudo-data at the beginning of training}, demonstrate that our method, \algopt, can achieve better performance with significantly less pseudo-data.
2) Pseudo-data can be used for various tasks.
Pseudo-data created using CIFAR10 performs well on tasks with local data from CIFAR10 and CIFAR100, as seen in Figure~\ref{Performance on different types of pseudo-data}.

\subsection{Construction of the Pseudo-Data} \label{sec:construction}
The choice of the pseudo-data in our \algopt framework is arbitrary.
For ease of presentation and taking the communication cost into account, we showcase two construction approaches below and detail their performance gain over all other existing baselines in Section~\ref{sec:experiments}:
\begin{itemize}[leftmargin=12pt,nosep]
	\item \textbf{Random Sample Mean (RSM)}.
	      Similar to the treatment in FedMix~\citep{yoon2021fedmix}, one RSM sample of the pseudo-data is estimated through a weighted combination of a random subset of local samples, and the pseudo-label is set\footnote{
		      We assume that pseudo-data does not belong to any particular classes, and should not give high confidence to any of that.
		      \looseness=-1
	      } to $\tilde{\mathbf{y}_p} = \frac{1}{C} \cdot \mathbf{1}$.
	      It is worth noting that RSM \textit{does not require the local data to be balanced} when constructing the pseudo-data, as long as the local data is distinct from the pseudo-data.
	      We show in Figure~\ref{Performance on different label distribution of RSM} that our algorithm (\algopt) can achieve comparable performance using pseudo-data constructed from data with unbalanced label distribution.
	      For more details, see Algorithm \ref{Construct Augmentation Data} in the appendix.
	\item \textbf{Mixture of local samples and the sample mean of a proxy dataset (Mixture)}.
	      This strategy relies on applying the procedure of RSM to irrelevant and globally shared proxy data (refer to Algorithm~\ref{Construct Augmentation Data by Proxy Data}).
	      To guard the distribution distance between the pseudo-data and local data, one sample of the pseudo-data at each client is constructed by
	      \begin{small}
		      \begin{align}
			      \textstyle
			      \tilde{\xx}_p \!=\! \frac{1}{K \!+\! 1} ( \xx_p \!+\! \sum_{k \!=\! 1}^{K} \xx_k ) ,
			      %   \qquad
			      \tilde{\yy}_p \!=\! \frac{1}{K \!+\! 1} ( \frac{1}{C} \!\cdot\! \mathbf{1} \!+\! \sum_{k \!=\! 1}^{K} \yy_k ) \,,
			      \label{pseudo_data}
		      \end{align}
	      \end{small}%
	      where $\xx_p$ is one RSM sample of the global proxy dataset, and $\xx_k$ and $\yy_k$ correspond to the data and label of one local sample (vary depending on the client).
	      $K$ is a constant that controls the closeness between the distribution of pseudo-data and local data.
	      As we will show in Section~\ref{sec:experiments}, setting $K = 1$ is data-efficient yet sufficient to achieve good results.
\end{itemize}

\paragraph{Remark: preserving privacy via Mixture.}
The RSM method is similar to the data augmentation method used in FedMix.
Similar to the discussion in FedMix, such a scheme may leak privacy.
To address this, we propose using the Mixture as a privacy-preserving method.
Mixture can even outperform RSM as justified in Figure~\ref{Performance on different types of pseudo-data}.

\begin{algorithm}[!t]
	\small
	\begin{algorithmic}[1]
		\small
		\Require{Local datasets $D_1, \dots, D_N$, pseudo dataset $D_{p}$ where $|D_p| = B$, and $B$ is the batch size,  number of local iterations $K$, number of communication rounds $T$, number of clients chosen in each round $M$, weights used in designed loss $\lambda, \mu$, local learning rate $\eta$.}
		\Ensure{Trained model $\momega_T$, $\boldsymbol{\theta}_T$, $\mphi_T$.}
		\myState{Initialize $\momega_0, \boldsymbol{\theta}_0, \mphi_0$.}
		\For{$t = 0, \dots, T-1$}
		\myState{Send $\momega_t, \boldsymbol{\theta}_t, \mphi_t$, $D_p$ (optional) to all clients.}
		\For{chosen client $i = 1, \dots, M$}
		\myState{$\momega_i^{0} = \momega_t, \boldsymbol{\theta}_i^{0} = \boldsymbol{\theta}_t, \mphi_i^{0} = \mphi_t, \mphi_g = \mphi_t$}
		\For{$k = 1, \dots, K$}
		\myState{\# Max Step}
		\myState{$\boldsymbol{\theta}_i^{k} = \boldsymbol{\theta}_i^{k-1} + \eta \nabla_{\boldsymbol{\theta}} \cL_{adv}$.}
		\myState{\# Min Step}
		\myState{$\momega_i^{k} = \momega_i^{k-1} - \eta \nabla_{\momega} \cL_{k}.$}
		\myState{$\mphi_i^{k} = \mphi_i^{k-1} - \eta \nabla_{\mphi} \cL_{gen}.$}
		\EndFor
		\myState{Send $\momega_i^{K}, \boldsymbol{\theta}_i^{K}, \mphi_i^{K}$ to server.}
		\EndFor
		\myState{$\momega_{t+1} = \frac{1}{M} \sum_{i=1}^{M} \momega_i^{K}$.}
		\myState{$\boldsymbol{\theta}_{t+1} = \frac{1}{M} \sum_{i=1}^{M} \boldsymbol{\theta}_i^{K}$.}
		\myState{$\mphi_{t+1} = \frac{1}{M} \sum_{i=1}^{M} \mphi_i^{K}$.}
		\EndFor
	\end{algorithmic}
	\mycaptionof{algorithm}{\small Algorithm Framework of \algopt}
	\label{Algorithm Framework of AugCA}
\end{algorithm}

\subsection{Component 1: Reducing Bias in Local Classifiers} \label{sec:component1}
Due to the issue of label distribution skew or the absence of some samples for the majority/minority classes, the trained local model classifier tends to overfit the locally presented classes, and may further hinder the quality of the feature extractor (as justified in Figure \ref{Class imbalance figure} and Proposition~\ref{bias caused by local updates}).
\looseness=-1

As a remedy, here we implicitly mimic the global data distribution---by using the pseudo-data constructed in Section~\ref{sec:construction}---to regularize the outputs and thus debias the classifier (note that Component 1 is the second term of~\eqref{classification step eqation}):
\looseness=-1
\begin{align*}
	\lambda \mathbb{E}_{\xx_p\sim D_i} \left[ \cL_{cls} (\xx_p, \tilde{\yy_p}, \mphi_i, \momega) \right] \,.
\end{align*}
\looseness=-1

\subsection{Component 2: Reducing Bias in Local Features}
\label{sec:component2}

% \begin{wrapfigure}{R}{0.61\textwidth}
% \begin{minipage}{0.51\textwidth}
% \begin{figure*}
% 	\centering
% 	\subfigure[Model structure and optimization flow of AugCA-O]{\includegraphics[width=0.75\textwidth]{figures/model1.png}
% 	\label{fig:model1}
% 	}
% 	\subfigure[Model structure and optimization flow of AugCA-S]{\includegraphics[width=0.75\textwidth]{figures/model2.png}
% 		\label{fig:model2}
% 	}
% 	\vspace{-1em}
% 	\caption{\small \textbf{Model structures and optimization flows of AugCA}. 
% 	    The sub-figure (a) and (b) describes AugCA-O and AugCA-S, respectively. In AugCA-O, we use the local and global feature of augmentation pseudo-data, with respect to $F_g(X_A)$ and $F_i(X_A)$ to calculate the contrastive loss. In AugCA-S, we use $F_i(X_A)$ together with the local feature of real data $F_i(X)$ for the contrastive loss. Both two methods use real local data for classification loss.
% 	    \looseness=-1
% 	   % In the first stage of local steps, projection layers are updated to maximize the contrastive loss, and in the second stage, local models are updated to minimize the classification loss and the contrastive loss. \tang{what's first stage and second stage? here are two subfigures..} \guo{may remove it}
%     }
%     \vspace{-1em}
% 	\label{Model structures of AugCA}
% \end{figure*}
% \end{wrapfigure}

% \end{minipage}

% \end{wrapfigure}

In addition to alleviating the biased local classifier in Section~\ref{sec:component1}, here we introduce a crucial adversary strategy to learn unbiased local features.
% To overcome the ``bias'' of features in Proposition~\ref{bias caused by local updates}, we propose AugCA to learn client invariant features.

\paragraph{Intuition of constructing an adversarial problem.}
As discussed in Proposition~\ref{bias caused by local updates}, effective federated learning on heterogeneous data requires learning debiased local feature extractors that 1) can extract local features that are close to global features of the same input data; 2) can extract different local features for input samples from different distributions.
However, existing methods that directly minimize the distance between global features and local features~\citep{lifederated2018,li2021model} have limited performance gain (c.f.\ Table~\ref{Performance of algorithms}) due to the diminishing optimization objective caused by the indistinguishability between the global and local features of the same input.
To this end, we propose to extend the idea of adversarial training to our FL scenarios:
\begin{enumerate}[leftmargin=12pt, nosep]
	\item We construct a projection layer as the critical step to distinguish features extracted by the global and local feature extractor: such layer ensures that the projected features extracted by the local feature extractor will be close to each other (even for distinct local data distributions), but the difference between features extracted by the global and local feature extractor after projection will be considerable (even for the same input samples).
	\item We can find that constructing such a projection layer can be achieved by maximizing the local feature bias discussed in Proposition~\ref{bias caused by local updates}.
	      More precisely, it can be achieved by maximizing the distance between global and local features of pseudo-data and simultaneously minimizing the distance between local features of pseudo-data and local data.
	\item We then minimize the local feature biases (discussed in Proposition~\ref{bias caused by local updates}) under the trained projection space, to enforce the learned local features of pseudo-data to be closer to the global features of pseudo-data but far away from the local features of real local data.
\end{enumerate}

\paragraph{On the importance of utilizing the projection layer to construct the adversary problem.}
To construct the aforementioned adversarial training strategy, we consider using an additional projection layer to map features onto the projection space\footnote{
	Such a projection layer is not part of the feature extractor or used for classification, as shown in Figure \ref{Overview of FedAug}.
}.
In contrast to the existing works that similarly add a projection layer~\citep{li2021model}, we show that
1) simply adding a projection layer as part of the feature extractor has trivial performance gain (c.f.\ Figure~\ref{Performance of algorithms with/without additional projection layer});
2) our design is the key step to reducing the feature bias and boosting the federated learning on heterogeneous data (c.f.\ Table \ref{ablation study}).
\looseness=-1

% however, we would like to show that this will not have a significant performance gain, as we shown in Figure~\ref{Accuracy of different algorithms on CIFAR10 with/without additional projection layer}.
% In our design, the projection layer will not be used as part of feature extractor or used for classification task -- as shown in Figure \ref{Overview of FedAug}, we use the projection layer as kind of discriminator that can distinguish global and local features. We show the necessity of using the adversarial projection in Table \ref{ablation study} of Section~\ref{sec:ablation study}---without the projection, there is no improvement for our proposed method.

\paragraph{Objective function design.}
% As illustrated in previous paragraph, we are willing to minimize the distance of global and local features on the same data, and distinguish local features of different data. 
We extend the idea of \citet{li2021model} and improve the contrastive loss initially proposed in simCLR~\citep{chen2020simple} to our challenging scenario.
Different from previous works, we use the projected features (global and local) on pseudo-data as the positive pairs and rely on the projected local feature of both pseudo-data and local data as the negative pairs:
\begin{small}
	\begin{align}
		\textstyle
		 & f_1 = \exp\left( \frac{\text{sim}\left( P_{\mtheta}(\mphi_i(\xx_p) ), P_{\mtheta}(\mphi_g(\xx_p)) \right)}{\tau_1} \right) \,, \\
		 & f_2 = \exp\left( \frac{\text{sim}\left( P_{\mtheta}(\mphi_i(\xx_p)), P_{\mtheta}(\mphi_i(\xx)) \right)}{\tau_2} \right) \,,    \\
		 & \cL_{con}(\xx_p, \xx, \mphi_g, \mphi_i, \mtheta) = - \log \left( \frac{ f_1 }{ f_1 + f_2} \right) \,,
		\label{contrastive loss}
	\end{align}
\end{small}%
% \tao{let's give $\exp\left( \frac{\text{sim}\left( P_{\mtheta}(\mphi_i(\xx_p) ), P_{\mtheta}(\mphi_g(\xx_p)) \right)}{\tau_1} \right)$ and $\exp\left( \frac{\text{sim}\left( P_{\mtheta}(\mphi_i(\xx_p)), P_{\mtheta}(\mphi_i(\xx)) \right)}{\tau_2} \right)$ two names, so we can simplify the above equation}
where $P_{\mtheta}$ is the projection layer parameterized by $\mtheta$, $\tau_1$ and $\tau_2$ are temperature parameters, and $\text{sim}$ is the cos-similarity function.
Our implementation uses a tied value for $\tau_1$ and $\tau_2$ for the sake of simplicity, but an improved performance may be observed by tuning these two.

\begin{figure}
	\centering
	% \vspace{-1em}
	% \subfigure[\small Convergence curve on RotatedMNIST]{\includegraphics[width=0.32\textwidth]{./figures/convergence-mnist.pdf}}
	\subfigure[\small CIFAR10 Convergence]{\includegraphics[width=0.23\textwidth]{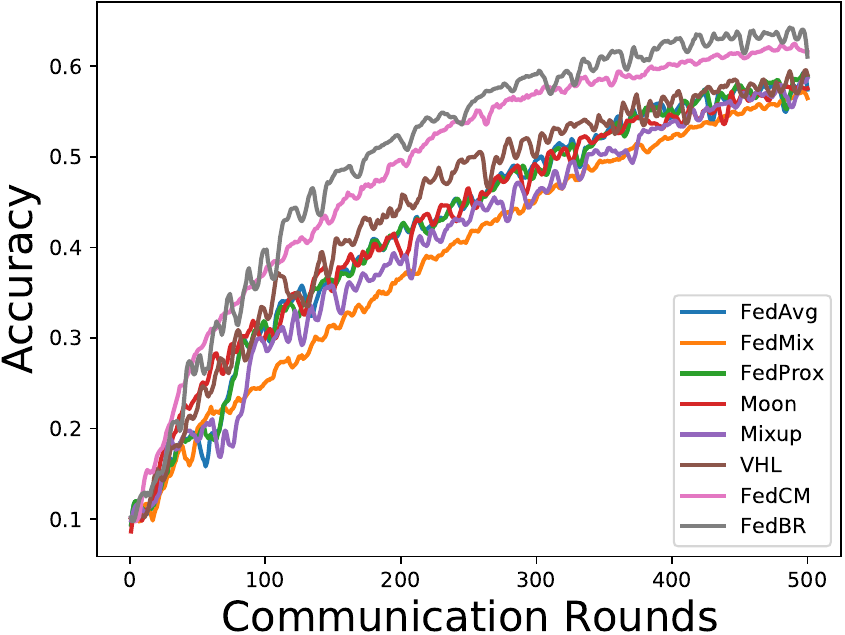}}
	\subfigure[\small CIFAR100 Convergence]{\includegraphics[width=0.23\textwidth]{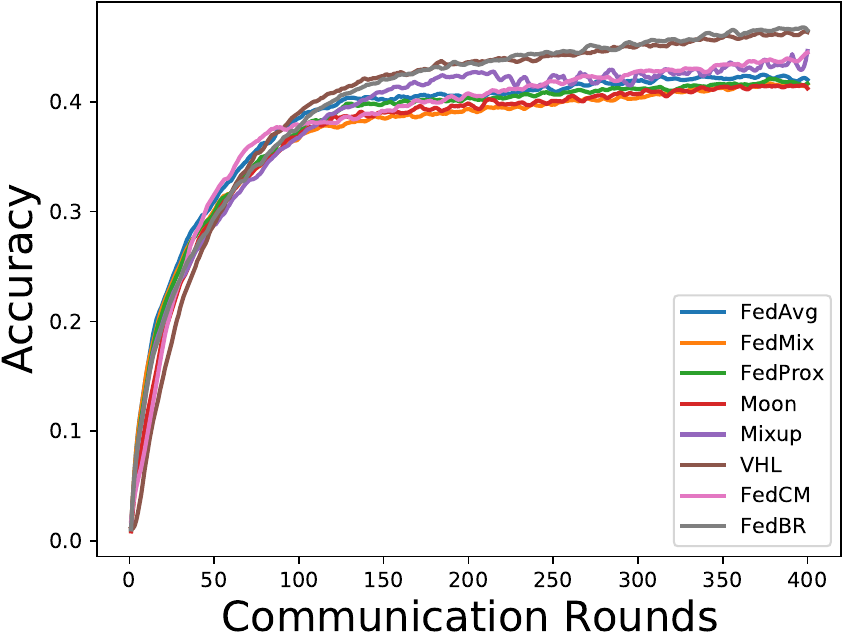}}
	\vspace{-1em}
	\caption{\small \textbf{Convergence curve of algorithms on different datasets.} We split RotatedMNIST, CIFAR10, and CIFAR100 datasets to 10 clients, and report the mean accuracy on all local test datasets for each communications rounds. More Details refer to Figure~\ref{convergence curve appendix} of Appendix~\ref{sec:Additional-Results}.}
	\vspace{-0.5em}
	\label{convergence curve main}
\end{figure}
% \tao{remove the white space of these three figures; this figure is not discussed in the main paper.}

\begin{table*}[!t]
	\small
	\centering
	% \vspace{-2em}
	\caption{\small
		\textbf{Performance of algorithms.}
		We split RotatedMNIST, CIFAR10, and CIFAR100 to 10 clients with $\alpha = 0.1$, and ran 1000 communication rounds on RotatedMNIST and CIFAR10 for each algorithm, 800 communication rounds CIFAR100.
		We report the mean of maximum (over rounds) 5 test accuracies and the number of communication rounds to reach the threshold accuracy.
		% Only very small part of combinations in Test 3 can be found in Train sets.
	}
	\vspace{-1em}
	\label{Performance of algorithms}
	\resizebox{.9\textwidth}{!}{%
		\begin{tabular}{l c c c c c c c c c c c c}
			\toprule
			\multirow{2}{*}{Algorithm} & \multicolumn{2}{c}{RotatedMNIST (CNN)} & \multicolumn{2}{c}{CIFAR10 (VGG11)} & \multicolumn{2}{c}{CIFAR100 (CCT)}                                                          \\
			\cmidrule(lr){2-3} \cmidrule(lr){4-5} \cmidrule(lr){6-7}
			                           & Acc (\%)                               & Rounds for 80\%                     & Acc (\%)                           & Rounds for 55\%     & Acc (\%)       & Rounds for 43\% \\
			\midrule
			Local                      & 14.67                                  & -
			                           & 10.00                                  & -                                   & 1.31                               & -                                                      \\
			FedAvg                     & 82.47                                  & 828 (1.0X)                          & 58.99                              & 736 (1.0X)          & 44.00          & 550 (1.0X)      \\
			FedProx                    & 82.32                                  & 824 (1.0X)                          & 59.14                              & 738 (1.0X)          & 43.09          & 756 (0.7X)      \\
			% 			SCAFFOLD        & 82.49 & 814 (1.0X) & 59.00 & 738 (1.0X) & 64.49 & 168 (1.0X) &\\ 
			Moon                       & 82.68                                  & 864 (0.9X)                          & 58.23                              & 820 (0.9X)          & 42.87          & 766 (0.7X)      \\
			DANN                       & 84.83                                  & 743 (1.1X)                          & 58.29                              & 782 (0.9X)          & 41.83          & -               \\
			GroupDRO                   & 80.23                                  & 910 (0.9X)                          & 56.57                              & 835 (0.9X)          & 44.34          & 444 (1.2X)      \\
			% FedCM & & & 62.63 & 508 (1.4X) & \\
			% FedNTD                     & 71.63                                  & -                                   & 55.69                              & 944 (0.8X)          & 45.21          & 352 (1.5X)      \\
			% FedDecorr & & & 54.15 & - &  \\
			% 			VHL             & 90.57      & 472 (1.8X)           & 60.04 & 652 (1.1X) & 44.04 & 284 (1.9X) \\
			% 			\midrule
			\algopt (Ours)             & \textbf{86.58}                         & \textbf{628 (1.3X)}                 & 64.65                              & 496 (1.5X)          & 45.14          & 352 (1.5X)      \\
			\midrule
			FedAvg + Mixup             & 82.56                                  & 840 (1.0X)                          & 58.57                              & 826 (0.9X)          & 46.37          & 358 (1.6X)      \\
			FedMix                     & 81.33                                  & 902 (0.9X)                          & 57.37                              & 872 (0.8X)          & 42.69          & -               \\
			% 			VHL + Mixup     & \textbf{91.56}      & \textbf{458 (1.8X)}           & 61.48 & 724 (1.0X)           & 46.66 & \textbf{266 (2.0X)} \\
			% 			\midrule
			\algopt + Mixup (Ours)     & 83.42                                  & 736 (1.1X)                          & \textbf{65.32}                     & \textbf{392 (1.9X)} & \textbf{47.75} & 294 (1.9X)      \\
			% MNIST train   & 81.22 & & 82.51 & 84.88 & 82.57 & 80.35 & 84.58 &
			\bottomrule
		\end{tabular}%
	}
	\vspace{-1.em}
\end{table*}

\begin{table}[!t]
	\small
	\centering
	% \vspace{-1em}
	\caption{\small
		\textbf{Comparison with VHL.}
		We split CIFAR10 and CIFAR100 to 10 clients with $\alpha = 0.1$, and
		% run 1000 communication rounds on CIFAR10 for each algorithm and 800 communication rounds on CIFAR100. We 
		report the mean of maximum (over rounds) 5 test accuracies and the number of communication rounds to reach the threshold accuracy. We set different numbers of virtual data to check the performance of VHL, and pseudo-data only transfer once in \algopt (32 pseudo-data). For CIFAR100, we choose Mixup as the backbone.
		\looseness=-1
		% Only very small part of combinations in Test 3 can be found in Train sets.
	}
	\vspace{-1em}
	\label{Comparison with VHL}
	\resizebox{0.5\textwidth}{!}{%
		\begin{tabular}{l c c c c c c c c c c c c}
			\toprule
			\multirow{2}{*}{Algorithm} & \multicolumn{2}{c}{CIFAR10 (VGG11)} & \multicolumn{2}{c}{CIFAR100 (CCT)}                                        \\
			\cmidrule(lr){2-3} \cmidrule(lr){4-5}
			                           & Acc (\%)                            & Rounds for 60\%                    & Acc (\%)       & Rounds for 46\%     \\
			\midrule
			% 			FedAvg    & 82.47 & 828 (1.0X) & 58.99 & 736 (1.0X) & 42.62 & 550 (1.0X) \\
			%             VHL (32 virtual data)              \\
			% 			VHL (2000 virtual data)             & 60.04 & 652 (1.1X) & 44.04 & 284 (1.9X) \\
			% 			\algopt (Ours)   & 64.65 & 496 (1.5X) & 43.85 & 352 (1.6X) \\
			% 			\midrule
			% 			FedAvg + Mixup           & 82.56 & 840 (1.0X) & 58.57 & 826 (0.9X) & 44.49 & 342 (1.6X) \\
			VHL (2000 virtual data)    & 61.23                               & 886 (1.0X)                         & 46.80          & 630 (1.0X)          \\
			VHL (20000 virtual data)   & 59.65                               & 998 (0.9X)                         & 46.51          & 714 (0.9X)          \\
			\algopt (32 pseudo-data)   & \textbf{64.61}                      & \textbf{530 (1.8X)}                & \textbf{47.67} & \textbf{554 (1.1X)} \\
			% MNIST train   & 81.22 & & 82.51 & 84.88 & 82.57 & 80.35 & 84.58 &
			\bottomrule
		\end{tabular}%64.61 & 46.27
	}
	% \vspace{-1.5em}
\end{table}

\begin{table}[!t]
	\small
	\centering
	% \vspace{-1em}
	\caption{\small
		\textbf{Combining \algopt with other baselines.}
		We split CIFAR10 and CIFAR100 to 10 clients with $\alpha = 0.1$, and
		% run 1000 communication rounds on CIFAR10 for each algorithm and 800 communication rounds on CIFAR100. We 
		report the mean of maximum (over rounds) 5 test accuracies. For \algopt, pseudo-data only transfer once (32 pseudo-data) using \textbf{Mixture}. For CIFAR100, we choose Mixup as the backbone.
		\looseness=-1
		% Only very small part of combinations in Test 3 can be found in Train sets.
	}
	\vspace{-1em}
	\label{Combining with other baselines}
	\resizebox{0.5\textwidth}{!}{%
		\begin{tabular}{l c c c c c c c c c c c c}
			\toprule
			\multirow{2}{*}{Algorithm} & \multicolumn{2}{c}{CIFAR10 (VGG11)} & \multicolumn{2}{c}{CIFAR100 (CCT)}                                        \\
			\cmidrule(lr){2-3} \cmidrule(lr){4-5}
			                           & w/o \algopt                         & + \algopt                          & w/o \algopt & + \algopt              \\
			\midrule
			FedAvg                     & 58.99                               & 64.66 (+5.67)                      & 46.37       & 47.98 (+1.61)          \\
			FedCM                      & 62.63                               & \textbf{65.32 (+2.69)}             & 46.15       & 46.95 (+0.80)          \\
			% FedNTD  & 59.10 & 744 (1.0X) &  \\
			FedDecorr                  & 54.15                               & 62.70 (+8.55)                      & 47.18       & \textbf{48.34 (+1.16)} \\
			FedNTD                     & 59.10                               & 59.26 (+0.16)                      & 47.02       & 47.18 (+0.16)          \\
			% MNIST train   & 81.22 & & 82.51 & 84.88 & 82.57 & 80.35 & 84.58 &
			\bottomrule
		\end{tabular}%64.61 & 46.27
	}
	% \vspace{-1.5em}
\end{table}

\section{Experiments} \label{sec:experiments}

\subsection{Experiment Setting}
We elaborate on experiment settings in Appendix \ref{sec:Experiment Details}.
\looseness=-1

\paragraph{Baseline algorithms.}
We compare \algopt with both SOTA FL baselines including FedAvg~\citep{mcmahan2016communication}, Moon~\citep{li2021model}, FedProx~\citep{lifederated2018}, VHL~\citep{tang2022virtual}, FedMix~\citep{yoon2021fedmix}, FedNTD~\citep{lee2022preservation}, FedCM~\citep{xu2021fedcm}, and FedDecorr~\citep{shi2022towards} which are most relevant to our proposed algorithms.
Similar to a very recent study in benchmarking FL~\citep{bai2023benchmarking}, we also contain domain generalization (DG) methods as baselines and check their performance under standard FL settings.
For DG baselines, we choose GroupDRO~\citep{sagawa2019distributionally}, Mixup~\citep{yan2020improve}, and DANN~\citep{ganin2015domain}. We also discuss other DG baselines in Appendix~\ref{sec:Experiment Details}.
% and commonly used domain generalization (DG) baselines that can be adapted to FL scenarios.
% Note that we do not consider domain generalization scenarios and include DG baselines to check if DG methods can benefit FL on non-iid clients.
% For FL baselines, we choose FedAvg~\citep{mcmahan2016communication}, Moon~\citep{li2021model}, FedProx~\citep{lifederated2018}, VHL~\citep{tang2022virtual}, and FedMix~\citep{yoon2021fedmix}, which are most relevant to our proposed algorithms. For DG baselines, we choose GroupDRO~\citep{sagawa2019distributionally}, Mixup~\citep{yan2020improve}, and DANN~\citep{ganin2015domain}.
Unless specially mentioned, all algorithms use FedAvg as the backbone algorithm.
\looseness=-1

\begin{figure}[!t]
	% \vspace{-2.4em}
	\centering
	\subfigure[\small w/ and w/o projection layer.]{\includegraphics[width=0.23\textwidth]{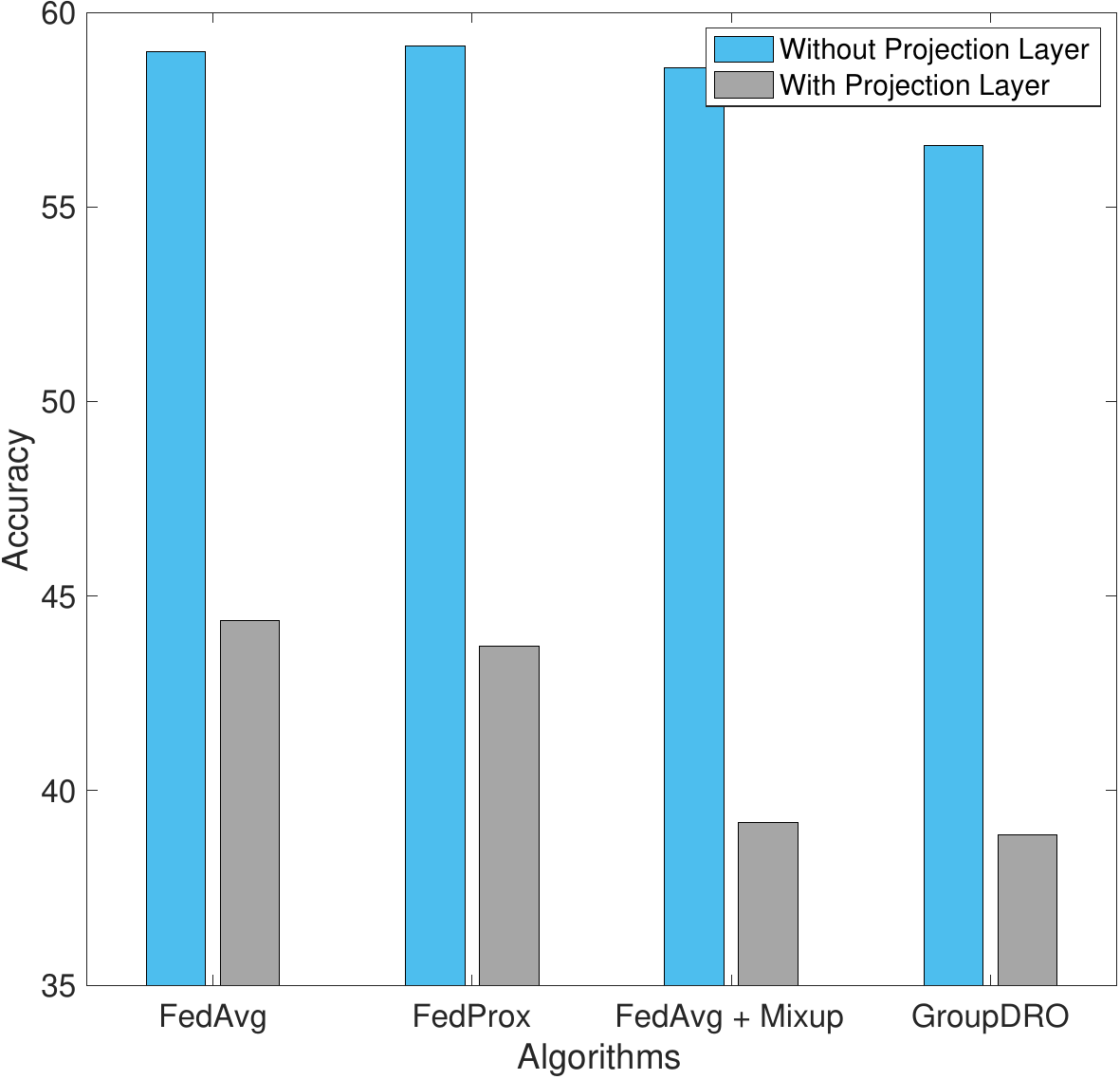}\label{Performance of algorithms with/without additional projection layer}}
	\subfigure[\small comm.\ of pseudo-data.]{\includegraphics[width=0.23\textwidth]{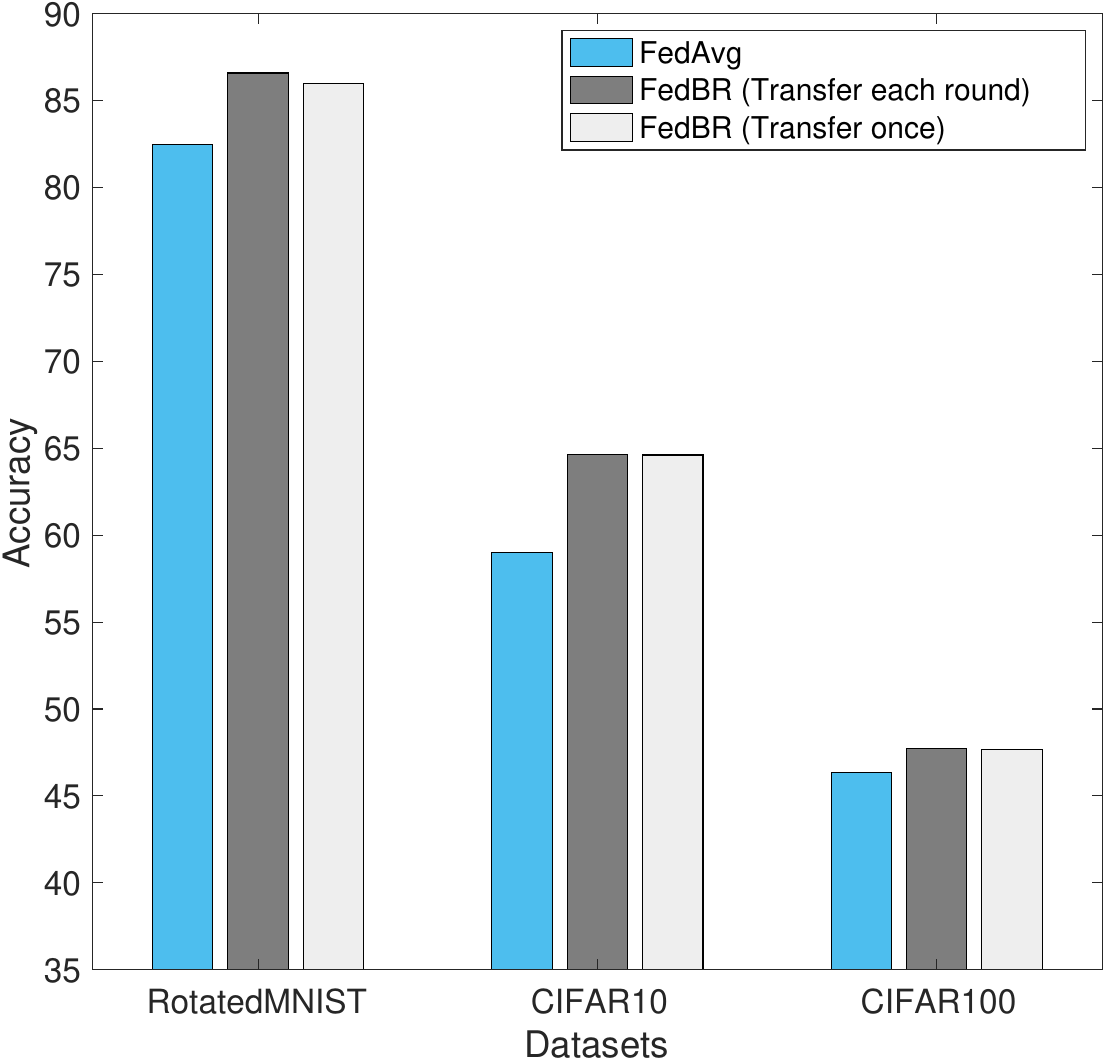}\label{Performance when only transfer pseudo-data at the beginning of training}} \\
	\vspace{-1em}
	\subfigure[\small choices of pseudo-data.]{\includegraphics[width=0.23\textwidth]{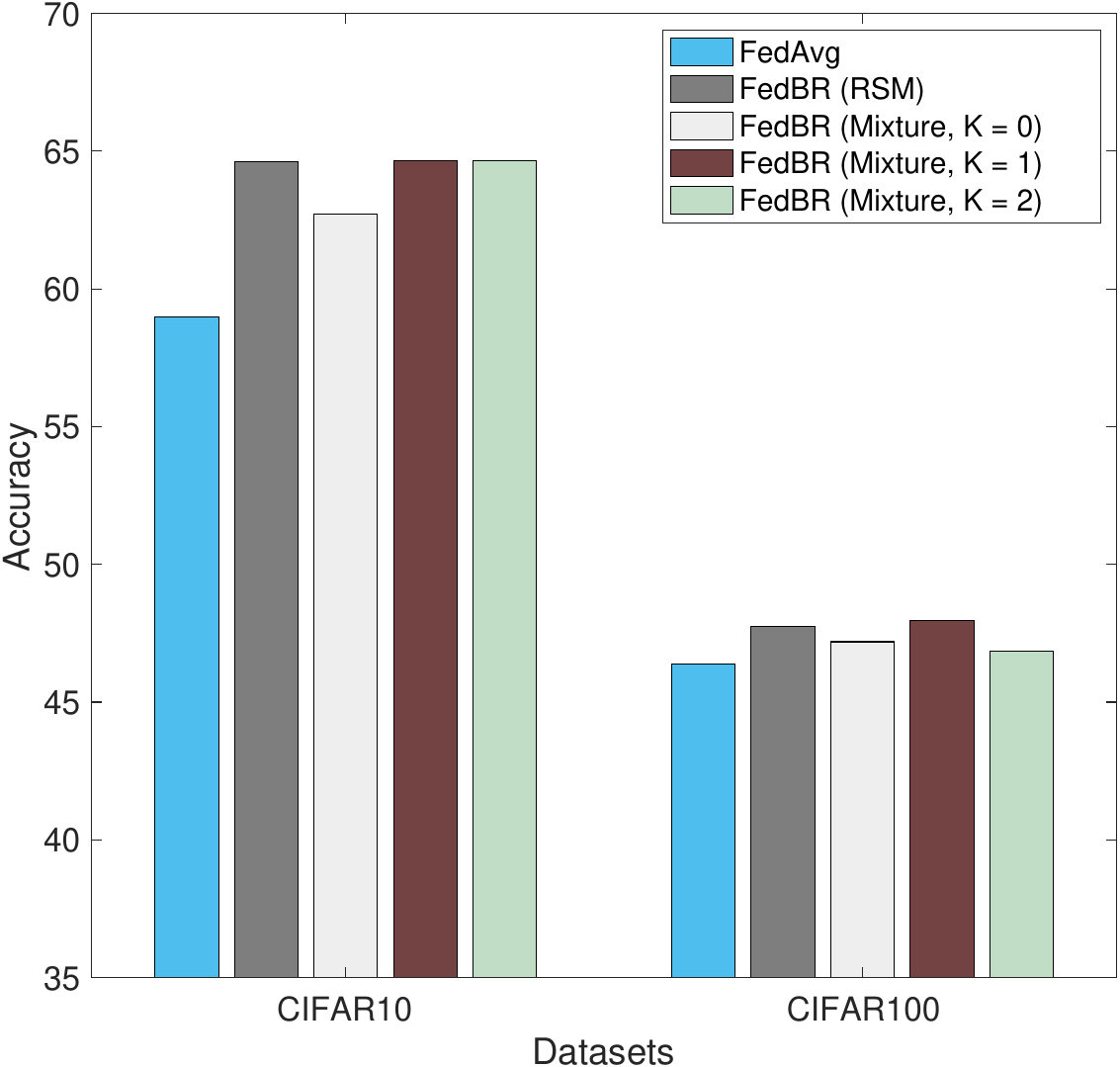}\label{Performance on different types of pseudo-data}}
	\subfigure[\small label distribution of RSM.]{\includegraphics[width=0.23\textwidth]{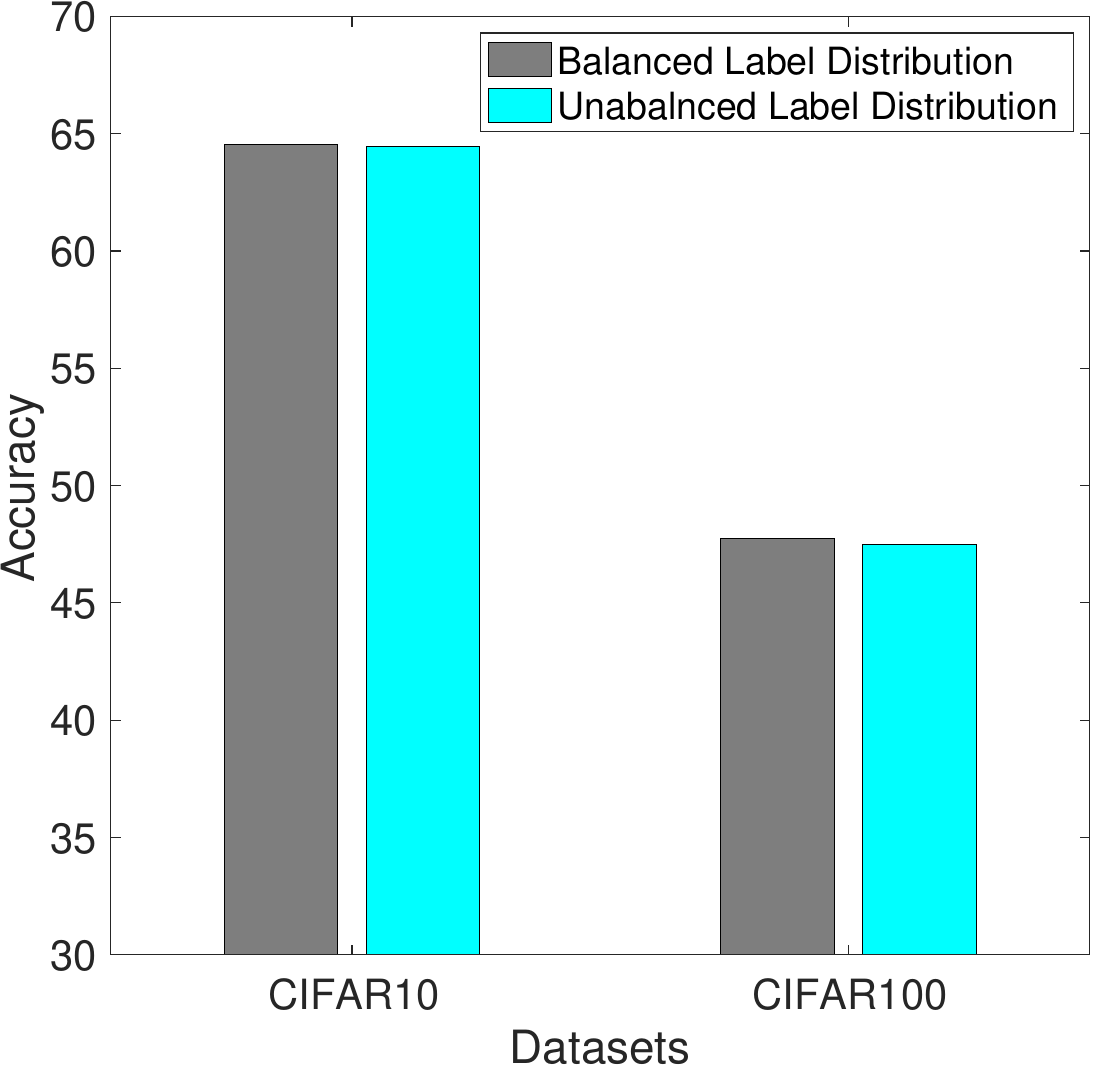}\label{Performance on different label distribution of RSM}}
	\vspace{-1em}
	\caption{\small
		\textbf{Ablation studies of \algopt}, regarding the impact of the projection layer, the communication strategy of pseudo-data, and the choices of pseudo-data.
		In Figure~\ref{Performance of algorithms with/without additional projection layer}, we show the performance of algorithms with/without the additional projection layer on the CIFAR10 dataset with the VGG11 model.
		In Figure~\ref{Performance when only transfer pseudo-data at the beginning of training}, we show the performance of \algopt on RotatedMNIST, CIFAR10, and CIFAR100 datasets when only transferring pseudo-data once (at the beginning of training) or generating new pseudo-data each round.
		In Figure~\ref{Performance on different types of pseudo-data}, we show the performance of \algopt using different types of pseudo-data.
		In Figure~\ref{Performance on different label distribution of RSM}, we show the performance of \algopt when constructing RSM using data with balanced and unbalanced label distribution.
		Pseudo-data \textit{transfer once} at the beginning of the training in  Figure~\ref{Performance on different types of pseudo-data}, and Figure~\ref{Performance on different label distribution of RSM}.
		% We split each dataset into 10 clients with $\alpha = 0.1$ and used CNN for RotatedMNIST dataset, VGG11 for CIFAR10, and CCT for CIFAR100. We run 1000 communication rounds on RotatedMNIST and CIFAR10 for each algorithm and 800 communication rounds on CIFAR100. We report the mean of maximum 5 test accuracies.
		\looseness=-1
	}
	\vspace{-1.5em}
	\label{Additional ablation studies}
\end{figure}

\paragraph{Models and datasets.} We examine all algorithms on RotatedMNIST, CIFAR10, and CIFAR100 datasets.
We use a four-layer CNN for RotatedMNIST, VGG11 for CIFAR10, and Compact Convolutional Transformer (CCT~\citep{hassani2021escaping}) for CIFAR100.
We split the datasets following the idea introduced in~\cite{yurochkin2019bayesian,hsu2019measuring,reddi2021adaptive}, where we leverage the Latent Dirichlet Allocation (LDA) to control the distribution drift with parameter $\alpha$.
The pseudo-data is chosen as \textbf{\textit{RSM}} by default, and we also provide results on other types of pseudo-data (c.f.\ Figure~\ref{Performance on different types of pseudo-data}).
By default, we generate one batch of pseudo-data (64 for MNIST and 32 for other datasets) in each round, and we also investigate only generating one batch of pseudo-data at the beginning of training to reduce the communication cost (c.f.\ Figure~\ref{Performance when only transfer pseudo-data at the beginning of training}, Figure~\ref{Performance on different types of pseudo-data}).
We use SGD optimizer and set the learning rate to $0.001$ for RotatedMNIST, and $0.01$ for other datasets.
The local batch size is set to 64 for RotatedMNIST, and 32 for other datasets (following the default setting in DomainBed ~\citep{gulrajani2020in}).
Additional results regarding the impact of hyper-parameter choices and performance gain of \algopt on other datasets/settings/evaluation metrics can be found in Appendix \ref{sec:Additional Results}.
\looseness=-1

\subsection{Numerical Results}
\paragraph*{The superior performance of \algopt over existing FL and DG algorithms.\footnote{
		See CIFAR10 + ResNet18 results in Table~\ref{Performance of cifar10 on resnet} of Appendix~\ref{sec:Additional Results}.}
	\looseness=-1
}
In Table \ref{Performance of algorithms} and Figure~\ref{convergence curve main}, we show the performance and convergence curve of baseline methods as well as our proposed \algopt algorithm.
When comparing different FL and DG algorithms, we discovered that:
1) \algopt performs best in all settings; 2) DG baselines only slightly outperform ERM, and some are even worse; 3) Regularizing local models to global models from prior rounds, such as Moon and Fedprox, does not result in positive outcomes.
% 4) Although the performance of global Mixup\tao{this term is confusing} idea is limited by the Mixup parameter $\lambda$. To make the Taylor Extension\tao{what's this? did you include the corresponding details?} work well, we must choose small $\lambda$, which entails a tinier improvement.
% \guo{maybe it's not necessary to detail comment the fedmix?}

\setcounter{subfigure}{0}

\begin{table*}[!t]
	% 	\vspace{-0.75em}
	\small
	\centering
	\caption{\small
		\textbf{Ablation studies of \algopt} on the effects of two components.
		We show the performance of two components and remove the max step (Line 8 in Algorithm~\ref{Algorithm Framework of AugCA}) of component 2.
		We split RotatedMNIST, CIFAR10, and CIFAR100 to 10 clients with $\alpha = 0.1$.
		We run 1000 communication rounds on RotatedMNIST and CIFAR10 for each algorithm and 800 communication rounds on CIFAR100.
		We report the mean of maximum (over rounds) 5 test accuracies and the number of communication rounds to reach the target accuracy.
		% Only very small part of combinations in Test 3 can be found in Train sets.
	}
	\vspace{-1em}
	\label{ablation study}
	\resizebox{.9\textwidth}{!}{%
		\begin{tabular}{l c c c c c c c c c c c c}
			\toprule
			\multirow{2}{*}{Algorithm} & \multicolumn{2}{c}{RotatedMNIST (CNN)} & \multicolumn{2}{c}{CIFAR10 (VGG11)} & \multicolumn{2}{c}{CIFAR100 (CCT)}                                                              \\
			\cmidrule(lr){2-3} \cmidrule(lr){4-5} \cmidrule(lr){6-7}
			                           & Acc (\%)                               & Rounds for 80\%                     & Acc (\%)                           & Rounds for 55\%     & Acc (\%)       & Rounds for 43\%     \\
			\midrule
			FedAvg                     & 82.47                                  & 828 (1.0X)                          & 58.99                              & 736 (1.0X)          & 46.37          & 358 (1.0X)          \\
			\midrule
			Component 1                & 84.40                                  & 770 (1.1X)                          & 64.32                              & \textbf{442 (1.7X)} & 47.22          & 330 (1.1X)          \\
			\, + min step              & 80.81                                  & 922 (0.9X)                          & 62.98                              & 562 (1.3X)          & 46.54          & 358 (1.0X)          \\
			Component 2                & 86.25                                  & 648 (1.3X)                          & 63.44                              & 483 (1.5X)          & \textbf{47.78} & 308 (1.2X)          \\
			\, + w/o max step          & 81.24                                  & 926 (0.9X)                          & 58.84                              & 584 (1.3X)          & 43.50          & 512 (0.7X)          \\
			\algopt                    & \textbf{86.58}                         & \textbf{628 (1.3X)}                 & \textbf{64.65}                     & 496 (1.5X)          & \textbf{47.75} & \textbf{294 (1.2X)} \\
			\bottomrule
		\end{tabular}%
	}
	% \vspace{-1.25em}
\end{table*}

\paragraph{Comparison with VHL.}
We vary the size of virtual data in VHL and compare it with our \algopt in Table~\ref{Comparison with VHL}~\footnote{The performance of \algopt is slightly different from results in Table~\ref{Performance of algorithms} because we only use 32 pseudo-data here to make a fair comparison with VHL.}:
our communication-efficient \algopt only uses 32 pseudo-data and transfers pseudo-data once, while the communication-intensive VHL~\citep{tang2022virtual} requires the size of virtual data to be proportional to the number of classes and uses at least 2{,}000 virtual data (the authors suggest 2{,}000 for CIFAR10 and 20{,}000 for CIFAR100 respectively in the released official code, and we use the default value of hyper-parameters and implementation provided by the authors).
We can find that 1) \algopt always outperforms VHL. 2) \algopt overcomes several shortcomings of VHL, e.g.\ the need for labeled virtual data and the large size of the virtual dataset.
% Note that  is communication-intensive, and the size of virtual dataset is proportional to the number of classes in the original dataset~\citep{tang2022virtual}.
\looseness=-1

\subsection{Ablation Studies}

\label{sec:ablation study}

\paragraph{Effectiveness of the different components in \algopt.}
In Table \ref{ablation study}, we show the improvements brought by different components of \algopt.
In order to highlight the importance of our two components, especially the max-step (c.f.\ Line 8 in Algorithm~\ref{Algorithm Framework of AugCA}) in component 2, we first consider two components of \algopt individually, followed by removing the max-step.
We find that:
1) Two components of \algopt have individual improvements compared with FedAvg, but the combined solution \algopt consistently achieves the best performance.
2) The projection layer is crucial.
After removing projection layers, Component 2 of \algopt performs even worse than FedAvg; such insights may also explain the limitations of Moon~\citep{li2021model}.
\looseness=-1

\paragraph{Performance of \algopt on CIFAR10 with different number of clients.}
In Table \ref{Performance of FedAug on CIFAR10 with different number of clients}, we increase the number of clients to 100, and $10$ clients are randomly chosen in each communication round.
We can find that \algopt consistently outperform other methods.

% \begin{table}[!t]
% 	\small
% 	\centering
% 	% \vspace{-0.5em}
% 	\caption{\small
% 		\textbf{Performance of \algopt on CIFAR10 with different number of clients.}
% 		We split CIFAR10 dataset into 10, 30, and 100 clients with $\alpha = 0.1$.
% 		We run 1000 communication rounds for each algorithm on the VGG11 model and report the mean of the maximal 5 accuracies (over rounds) during training on test datasets.
% 		\looseness=-1
% 		% Only very small part of combinations in Test 3 can be found in Train sets.
% 	}
% 	\vspace{-1em}
% 	\resizebox{.5\textwidth}{!}{%
% 		\begin{tabular}{l c c c c c c c c c c}
% 			\toprule

% 			Methods & Acc (\%) with 10 clients & Acc (\%) with 30 clients & Acc (\%) with 100 clients \\
% 			\midrule
% 			FedAvg  & 58.99                    & 44.83                    & 38.20                     \\
% 			\algopt & \textbf{64.65}           & \textbf{50.28}           & \textbf{41.59}            \\
% 			% MNIST train   & 81.22 & & 82.51 & 84.88 & 82.57 & 80.35 & 84.58 &
% 			\bottomrule
% 		\end{tabular}%
% 	}
% 	\label{Performance of FedAug on CIFAR10 with different number of clients}
% 	\vspace{-1.25em}
% \end{table}

\begin{table}[!t]
	\small
	\centering
	% \vspace{-0.5em}
	\caption{\small
		\textbf{Performance of algorithms with 100 clients.}
		We split CIFAR10 dataset into 100 clients with $\alpha = 0.1$.
		We run 1000 communication rounds for each algorithm on the VGG11 model and report the mean of the maximal 5 accuracies (over rounds) during training on test datasets.
		\looseness=-1
		% Only very small part of combinations in Test 3 can be found in Train sets.
	}
	\vspace{-1em}
	\resizebox{.48\textwidth}{!}{%
		\begin{tabular}{l c c c c c c c c c c}
			\toprule

			Methods & FedAvg & FedDecorr & FedMix & FedProx & Mixup & VHL   & FedBR \\
			\midrule
			Acc     & 38.20  & 35.53     & 34.71  & 37.90   & 36.63 & 40.93 & 41.59 \\
			% Acc  & 58.99                    & 44.83                    & 38.20                     \\
			% \algopt & \textbf{64.65}           & \textbf{50.28}           & \textbf{41.59}            \\
			% MNIST train   & 81.22 & & 82.51 & 84.88 & 82.57 & 80.35 & 84.58 &
			\bottomrule
		\end{tabular}%
	}
	\label{Performance of FedAug on CIFAR10 with different number of clients}
	% \vspace{-1.25em}
\end{table}

\begin{table}[!t]
	\small
	\centering
	% \vspace{-1em}
	\caption{\small
		\textbf{Performance of local model on balanced global test datasets.}
		We split CIFAR10 to 10 clients with $\alpha = 0.1$, and
		% run 1000 communication rounds on CIFAR10 for each algorithm and 800 communication rounds on CIFAR100. We 
		report the test accuracies achieved by the local models/aggregated models at the end of each communication round. For \algopt, pseudo-data only transfer once (32 pseudo-data).
		\looseness=-1
		% Only very small part of combinations in Test 3 can be found in Train sets.
	}
	\vspace{-1em}
	\label{Performance of local model on balanced global test datasets}
	\resizebox{0.48\textwidth}{!}{%
		\begin{tabular}{l c c c c c c c c c c c c}
			\toprule
			Algorithm                    & FedAvg & FedDecorr & VHL   & \algopt \\
			\midrule
			Local Model Performance      & 21.01  & 21.18     & 32.81 & 21.83   \\
			Aggregated Model Performance & 46.37  & 47.10     & 46.80 & 47.67   \\
			% FedNTD  & 59.10 & 744 (1.0X) &  \\
			\bottomrule
		\end{tabular}%64.61 & 46.27
	}
	% \vspace{-1.5em}
\end{table}

\begin{table}[!t]
	\small
	\centering
	% \vspace{-1em}
	\caption{\small
		\textbf{Parameter transmitted and mean simulation time in each round.}
		We split CIFAR10 and CIFAR100 to 10 clients with $\alpha = 0.1$. For \algopt, pseudo-data only transfer once (32 pseudo-data). The simulation time only includes the computation time per step, and do not includes the communication time. CIFAR100 experiments use Mixup as backbone.
		\looseness=-1
		% Only very small part of combinations in Test 3 can be found in Train sets.
	}
	\vspace{-1em}
	\label{Parameter transmitted and mean simulation time in each round}
	\resizebox{0.48\textwidth}{!}{%
		\begin{tabular}{l c c c c c c c c c c c c}
			\toprule
			CIFAR10 (VGG11)          & FedAvg & Moon & VHL  & FedCM & \algopt \\
			\midrule
			Parameters (Millions)    & 9.2    & 9.7  & 9.2  & 18.4  & 9.7     \\
			Mean simulation time (s) & 0.29   & 0.69 & 0.43 & 0.36  & 0.60    \\
			\toprule
			CIFAR100 (CCT)           & FedAvg & Moon & VHL  & FedCM & \algopt \\
			\midrule
			Parameters (Millions)    & 22.4   & 22.6 & 22.4 & 44.8  & 22.6    \\
			Mean simulation time (s) & 0.67   & 1.97 & 1.44 & 0.85  & 1.19    \\
			% FedNTD  & 59.10 & 744 (1.0X) &  \\
			\bottomrule
		\end{tabular}%64.61 & 46.27
	}
	% \vspace{-1.5em}
\end{table}

\begin{figure}[!t]
	% \vspace{-2.4em}
	\centering
	\subfigure[CIFAR10]{\includegraphics[width=0.23\textwidth]{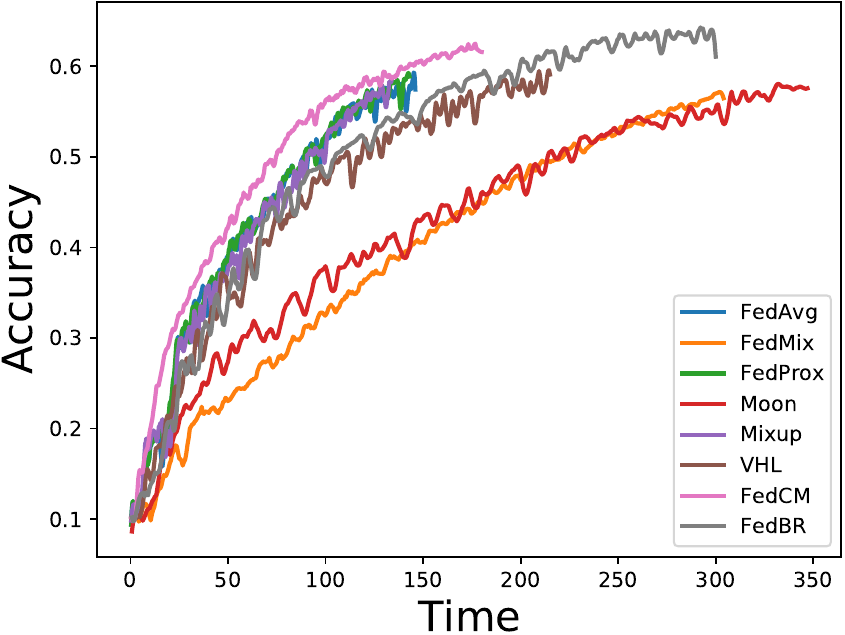}\label{fig:CIFAR10-time-smooth}}
	\subfigure[CIFAR100]{\includegraphics[width=0.23\textwidth]{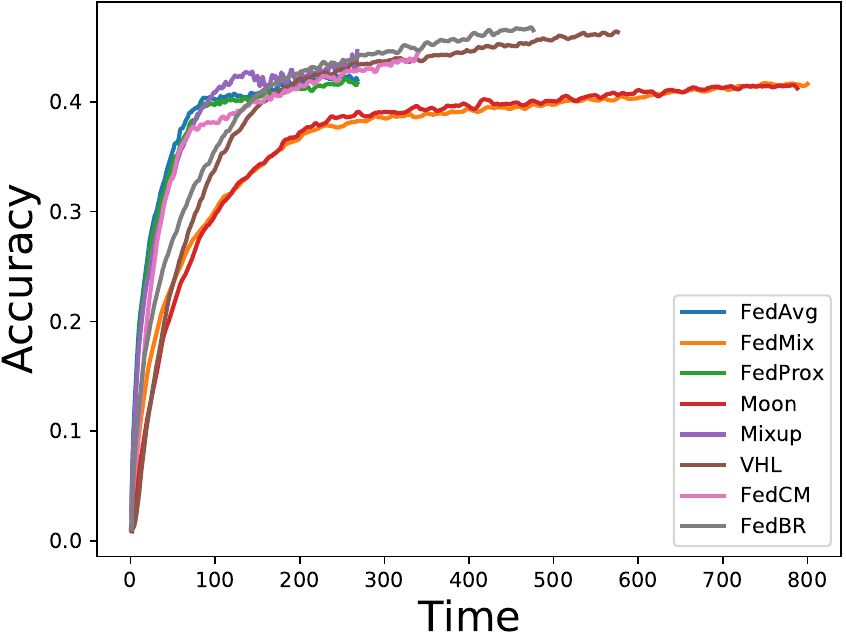}\label{fig:CIFAR100-time-smooth}} \\
	\vspace{-1em}
	\caption{\small
		\textbf{Convergence curve w.r.t. simulation time.} We split CIFAR10 and CIFAR100 datasets to 10 clients, and report the mean accuracy on all local test datasets at each time slots. We use VGG11 for CIFAR10 experiments, and CCT for CIFAR100 experiments. CIFAR100 experiments use Mixup as backbone for all the algorithms.
		More details refer to Figure~\ref{fig:Convergence curve w.r.t simulation time} of Appendix~\ref{sec:Additional-Results}.
		\looseness=-1
	}
	% 	\vspace{-1.5em}
	\label{Convergence curve w.r.t. simulation time}
\end{figure}

% Directly use \textbf{Mixture} is not good. However, setting $K = 1$ on \textbf{Mixture} (Equation~\eqref{pseudo_data}) is enough to achieve comparable performance with using \textbf{RSM}.\tao{logic here is strange and unclear}
\paragraph{Reducing the communication cost of \algopt.}
To reduce the communication overhead, we reduce the size of pseudo-data, and only transmit one mini-batch of pseudo-data (64 for MNIST and 32 for others) once at the beginning of training.
In Figure~\ref{Performance when only transfer pseudo-data at the beginning of training}, we show the performance of \algopt when pseudo-data only transfer to clients at the beginning of the training (64 pseudo-data for RotatedMNIST, and 32 for CIFAR10 and CIFAR100).
Results show that only transferring pseudo-data once can achieve comparable performance gain compared with transferring pseudo-data in each round.
This indicates that the performance of \algopt will not drop even if we give a small number of pseudo-data.
\looseness=-1

\paragraph{Regarding privacy issues caused by RSM.} Because RSM may have some privacy issues, we consider using Mixture to protect privacy.
In Figure~\ref{Performance on different types of pseudo-data}, we show the performance of \algopt with different types of pseudo-data (pseudo-data only transfer once at the beginning of training as in Figure~\ref{Performance when only transfer pseudo-data at the beginning of training}).
Results show that:
1) \algopt consistently outperforms FedAvg on all types of pseudo-data.
2) When using \textbf{Mixture} as pseudo-data and setting $K = 0$ (\eqref{pseudo_data}), \algopt still have a performance gain compared with FedAvg, and a more significant performance gain can be observed by setting $K = 1$.

\paragraph{Constructing pseudo-data by RSM using local data with unbalanced label distribution.} In Figure~\ref{Performance on different label distribution of RSM}, we construct the pseudo-data for \algopt using data with (1) balanced and (2) unbalanced label distributions. Results show that the performance of \algopt remained the same even when the data used to create the pseudo-data had an unbalanced label distribution.

\paragraph{Combining \algopt with other FL methods enhances performance.} In Table~\ref{Combining with other baselines}, we combine \algopt with other SOTA FL algorithms, including FedNTD~\citet{lee2022preservation}, FedCM~\citep{xu2021fedcm} and FedDecorr~\citep{shi2022towards}. Results demonstrate that \algopt significantly enhances the performance of these methods through simple integration.

\paragraph{Effectiveness of \algopt on reducing the local learning bias.} To validate \algopt's ability to train unbiased local models, we save and assess the local models at the end of each communication round using balanced global test datasets. Results in Table~\ref{Performance of local model on balanced global test datasets} show that: 1) FedBR can achieve better performance than FedAvg without using the labeled global shared data, and the aggregated model matches and even surpasses VHL's performance 2) Local models of VHL perform better than other methods by using labeled global shared datasets to correct classification errors. It is natural for VHL to achieve better local performance as local datasets of VHL is relatively balanced.

\paragraph{Performance of \algopt regarding communication and computation costs.} Using pseudo-data and an additional projection layer in \algopt increases computation and communication costs. We quantify this by reporting transmitted parameters and mean simulation time per round in Table~\ref{Parameter transmitted and mean simulation time in each round}, and display the convergence curve with respect to the simulation time in Figure~\ref{Convergence curve w.r.t. simulation time}. Results show that: 1) The computation time of \algopt is similar to that of other FL methods that adding regularization terms to overcome the local learning bias, such as VHL and Moon. 2) \algopt's communication cost remains minimal as it only introduce an additional small three-layer MLP projection layer, in contrast to the larger feature extractors found in modern deep neural networks.

% \begin{table*}[!ht]
% 	\small
% 	\centering
% % 	\vspace{-1em}
% 	\caption{\small
% 		\textbf{Performance of \algopt when augmentation data only transfer once.}
% 		We split RotatedMNIST and CIFAR10 datasets to 10 clients with $\alpha = 0.1$. We run 1000 communication rounds for each algorithm, and report the mean of maximum 5 accuracies during training on test datasets.
% 		% Only very small part of combinations in Test 3 can be found in Train sets.
% 	}
% 	\vspace{-0.75em}
% 	\label{Performance of FedAug when augmentation data only transfer once}
% 	\resizebox{1.\textwidth}{!}{%
% 		\begin{tabular}{l c c c c c c c c c c}
% 			\toprule

% 			Methods &  RotatedMNIST (CNN) & Acc (\%) on CIFAR10 (VGG11) & Acc (\%) on CIFAR100 (CCT) \\ 
% 			\midrule
% 			FedAvg & 82.47 & 58.99 & 44.49 \\
% 			\algopt (Transfer each round) & 86.58 & 64.65 & 46.87 \\
% 			\algopt (Transfer once) & 85.99 & 64.61 & 46.27 \\
% 			% MNIST train   & 81.22 & & 82.51 & 84.88 & 82.57 & 80.35 & 84.58 &
% 			\bottomrule
% 		\end{tabular}%
%  	}
%  	\vspace{-.75em}
% \end{table*}

\section{Conclusion and Future works}
We propose a new algorithm, \algopt, for Federated Learning that uses label-agnostic pseudo-data to improve performance on heterogeneous data. It has two key components and experiments show it significantly improves Federated Learning.
Unlike previous methods, \algopt does not require labeled pseudo-data or a large pseudo-dataset, therefore reducing the communication costs.

However, \algopt requires additional computation as the algorithm needs additional forward propagation on pseudo-data, as well as the additional computation on the min-max optimization procedure. It would be interesting to explore ways to reduce this extra computation in the future.
% In this paper, we present a new algorithm called \algopt that uses label-agnostic pseudo-data to debias local features and classifiers simultaneously in Federated Learning.
% \algopt has two key components that work together to improve performance on heterogeneous data.
% We conducted experiments to verify the effectiveness of \algopt and found that it significantly improves Federated Learning.
% Our proposed algorithm addresses several limitations of previous approaches, such as the need for labeled pseudo-data or a large pseudo-dataset.
% However, like other methods, it requires additional computation when training on pseudo-data.
% It would be interesting to explore ways to reduce this extra computation in the future.
% In this paper, we propose \algopt to reduce learning bias on local features and classifiers. The two components: AugMean and AugCA, complement each other to improve Federated Learning on heterogeneous data. Especially for AugCA, we propose to use projection space to distinguish global and local features, and and make the performance substantially better than other methods with similar intuition. Besides, 
% We verified the proposed methods

% \clearpage

\section*{Acknowledgement}
This work was supported in part by the National Key R\&D Program of China (Project No.\ 2022ZD0115100), the Research Center for Industries of the Future (RCIF) at Westlake University, and Westlake Education Foundation.
This work is also supported in part by the National Natural Science Foundation of China (Grant No.\ 72171206,
No.\ 71931003, No.\ 72061147004, No.\ 72192805 and No.\ 62001412), the National Key R\&D Program of China (Grant No.\ 2018YFB1800800), the Guangdong Provincial Key Laboratory of Future Networks of Intelligence (Grant No.\ 2022B1212010001), and the Shenzhen Institute of Artificial Intelligence and Robotics for Society (AIRS).

\bibliography{reference}
\bibliographystyle{common/icml2023}
%%%%%%%%%%%%%%%%%%%%%%%%%%%%%%%%%%%%%%%%%%%%%%%%%%%%%%%%%%%%
% \clearpage
% \input{checklist.tex}
%%%%%%%%%%%%%%%%%%%%%%%%%%%%%%%%%%%%%%%%%%%%%%%%%%%%%%%%%%%%
\clearpage
\appendix
% !TeX root = icml2023_fl_invariant.tex

\onecolumn
{
	\hypersetup{linkcolor=black}
	\parskip=0em
	\renewcommand{\contentsname}{Contents of Appendix}
	\tableofcontents
	\addtocontents{toc}{\protect\setcounter{tocdepth}{3}}
}

\section{Experiment Details}

\label{sec:Experiment Details}

\paragraph{Framework and baseline algorithms.} In addition to traditional FL methods, we aim to see if domain generalization (DG) methods can help increase model performance during FL training.
Thus, we use the DomainBed benchmark \citep{gulrajani2020in}, which contains a series of regularly used DG algorithms and datasets. The algorithms in DomainBed can be divided into three categories:

\begin{itemize}[leftmargin=12pt,nosep]
	\item \textbf{Infeasible methods:} Some algorithms can't be applied in FL scenarios due to the privacy concerns, for example, MLDG \citep{li2017learning}, MMD \citep{li2018domain}, CORAL \citep{sun2016deep}, VREx \citep{krueger2020out} that need features or data from each domain in each iteration.
	\item \textbf{Feasible methods (with limitations):} Some algorithms can be applied in FL scenarios with some limitations. For example, DANN \citep{ganin2015domain}, CDANN \citep{li2018deep} require knowing the number of domains/clients, which is impractical in the cross-device setting.
	\item \textbf{Feasible methods ( without limitations):} Some algorithms can be directly applied in FL settings. For example, ERM, GroupDRO \citep{sagawa2019distributionally}, Mixup \citep{yan2020improve}, and IRM \citep{martin2019invariant}.
\end{itemize}

We choose several common used DG algorithms that can easily be applied in Fl scenarios, including ERM, GroupDRO \citep{sagawa2019distributionally}, Mixup \citep{yan2020improve}, and DANN \citep{ganin2015domain}.
% We also summarize other OOD algorithms in Table~\ref{tab:} of Appendix \ref{}.
For FL baselines, we choose FedAvg~\citep{mcmahan2016communication} (equal to ERM), Moon~\citep{li2021model}, FedProx~\citep{lifederated2018}, SCAFFOLD~\citep{karimireddyscaffold2019} and FedMix~\citep{yoon2021fedmix} which are most related to our proposed algorithms.

Notice that some existing works consider combining FL and domain generalization. For example, combining DRO with FL~\citep{mohri2019agnostic,deng2021distributionally}, and combine MMD or DANN with FL~\citep{peng2019federated,wang2022framework,shen2021fedmm}. The natural idea of the former two DRO-based approaches is the same as our GroupDRO implementations, with some minor weight updates differences; the target of the later series of works that combine MMD or DANN is to train models to work well on unseen distributions, which is orthogonal with our consideration (overcome the local heterogeneity).To check the performance of this series of works, we choose to integrate FL and DANN into our environments.

Notice that we carefully tune all the baseline methods. The implementation detail of each algorithm is listed below:
\begin{itemize} [leftmargin=12pt,nosep]
	\item GroupDRO: The weight of each client is updated by $\momega_i^{t+1} = \momega_i^{t} \exp (0.01 l_i^{t})$, where $l_i^{t}$ is the loss value of client $i$ at round $t$.
	\item Mixup: Local data is mixed by $\tilde{\xx} = \lambda \xx_i + (1 - \lambda) \xx_j$, and $\lambda$ is sampled by $Beta(0.2, 0.2)$.
	\item DANN: Use a three-layer MLP as domain discriminator, where the width of MLP is 256. The weight of domain discriminate loss is tuned in $\{ 0.01, 0.1, 1 \}$.
	\item FedProx: The weight of proximal term is tuned in  $\{ 0.001, 0.01, 0.1 \}$.
	\item Moon: The projection layer is a two-layer MLP, the MLP width is setting to 256, and the output dimension is 128. We tuned the weight of contrastive loss in  $\{ 0.01, 0.1, 1, 10 \}$.
	\item FedMix: The mixup weight $\lambda$ used in FedMix is tuned in $\{ 0.01, 0.1, 0.2 \}$, we construct 64 augmentation data in each local step for RotatedMNIST, and 32 samples for CIFAR10 and CIFAR100..
	\item VHL: We use the same setting as in the original paper, with the weight of augmentation classification loss $\alpha = 1.0$, and use the "proxy\_align\_loss" provided by the authors for feature alignment. Virtual data is generated by untrained style-GAN-v2, and we sample 2000 virtual data for CIFAR10 and RotatedMNIST; 20000 virtual data for CIFAR100 follow the default setting of the original work. To make a fair comparison, we sample 32 virtual samples in each local step for CIFAR10 and CIFAR100.
        \item FedNTD: We use the official code of FedNTD, set $\tau = 1.0$ as suggested in the original paper, and $\beta$ is tuned in  $\{1.0, 0.1\}$
        \item FedDecorr: We use the official code of FedDecorr, and set the weight of penalty term to $0.1$.
	\item \algopt: We use a three-layer MLP as the projection layer, the MLP width is set to 256, and the output dimension is 128. By default, we set $\tau_1 = \tau_2 = 2.0$, the weight of contrastive loss $\mu = 0.5$, and the weight of AugMean $\lambda = 1.0$ on MNIST and CIFAR100, $\lambda = 0.1$ on CIFAR10 and PACS. We sample 64 pseudo-data in each local step for RotatedMNIST and 32 samples for CIFAR10 and CIFAR100.
\end{itemize}
% We choose to implement DANN in our environment as a sanity check.

% \paragraph{Feature correction when using proxy datasets to construct pseudo-data.} When using proxy datasets to construct the pseudo-data, we additionally mix up local data with pseudo-data to make the pseudo-data not too far from the local distribution. However, the pseudo-data will have a large overlap with local data after the mixup. Then the $\exp \left( \frac{\text{sim}(P(\mphi_i(x_p)), P(\mphi_i(x)))}{\tau_2} \right)$ term in Equation \eqref{contrastive loss}, which is used to maximize the distance between local features of local data and pseudo-data, will be meaningless.
% To address this issue, we change this term to
% \begin{align}
% 	\exp \left( \frac{\text{sim}\left( P(\mphi_i(\xx_p) - \left\langle \tilde{\yy_p}, \yy \right\rangle \cdot \mphi_i(\xx)), P(\mphi_i(\xx)) \right)}{\tau_2} \right) \,,
% \end{align}
% where $\tilde{y_p}$ is the pseudo-label of $x_p$, and $y$ is the one-hot label of local data $x$. Then we can minimize the relationship between $x$ and $x_p$ caused by the mixup with local data.

\paragraph{Datasets and Models.} For datasets, we choose RotatedMNIST, CIFAR10, CIFAR100, and PACS.
For RotatedMNIST, CIFAR10, and CIFAR100, we split the datasets following the idea introduced in~\cite{yurochkin2019bayesian,hsu2019measuring,reddi2021adaptive}, where we leverage the Latent Dirichlet Allocation (LDA) to control the distribution drift with parameter $\alpha$. Larger $\alpha$ indicates smaller non-iidness. We divided each environment into two clients for PACS, with the first client containing data from classes 0-3, and the second client containing data from classes 4-6.

Unless specially mentioned, we split RotatedMNIST, CIFAR10, and CIFAR100 to 10 clients and set $\alpha = 0.1$. For PACS, we have 8 clients instead. Notice that for each client of CIFAR10, we utilize a special transformation, i.e., rotation to the local data, to simulate the natural shift. In detail:

\begin{itemize} [leftmargin=12pt,nosep]
	\item RotatedMNIST: We first split MNIST by LDA using parameter $\alpha = 0.1$ to 10 clients, then for each client, we rotate the local data by $\{0, 15, 30, 45, 60, 75, 90, 105, 120, 135\}$.
	\item CIFAR10: We first split CIFAR10 by LDA using parameter $\alpha = 0.1$ to $N$ clients. Then for each client, we sample $q \in \mathbb{R}^{10}$ from $Dir(1.0)$. For each image in local data, we sample an angle in $\{0, 15, 30, 45, 60, 75, 90, 105, 120, 135\}$ by probability $q$, and rotate the image by the angle.
	\item Clean CIFAR10: Unlike the previous setting, we do not rotate the samples in CIFAR10 (no inner-class non-iidness).
	\item CIFAR100: We split the CIFAR100 by LDA using parameter $\alpha = 0.1$, and transform the train data using RandomCrop, RandomHorizontalFlip, and normalization.
\end{itemize}

Each communication round includes 50 local iterations, with 1000 communication rounds for RotatedMNIST and CIFAR10, 800 communication rounds for CIFAR100, and 400 communication rounds for PACS. Notice that the number of communication rounds is carefully chosen, and the accuracy of all algorithms does not significantly improve after the given communication rounds.

The public data is chosen as RSM~\citep{yoon2021fedmix} by default, and we also provide results on other proxy datasets.
We utilize a four-layer CNN for MNIST, VGG11 for CIFAR10 and PACS, and CCT~\citep{hassani2021escaping} (Compact Convolutional Transformer, cct\_7\_3x1\_32\_c100) for CIFAR100.

For each algorithm and dataset, we employ SGD as the optimizer, and set learning rate $lr = 0.001$ for MNIST, and $lr = 0.01$ for CIFAR10 , CIFAR100, and PACS. When using CCT and ResNet, we set momentum as $0.9$. We set the same random seeds for all algorithms.
We set local batch size to 64 for RotatedMNIST, and 32 for CIFAR10, CIFAR100, and PACS.
% The learning rate is set to $0.001$ for RotatedMNIST, and $0.01$ for others.

\section{Details of Augmentation Data}

\label{sec:details of augmentation data}

\begin{figure*}
	\centering
	\includegraphics[width=0.18\textwidth]{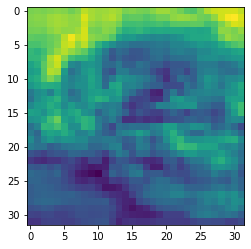}
	\includegraphics[width=0.18\textwidth]{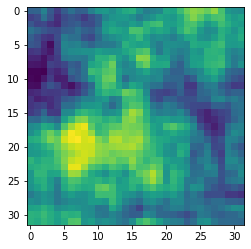}
	\includegraphics[width=0.18\textwidth]{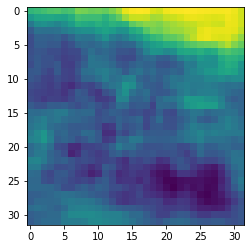}
	\includegraphics[width=0.18\textwidth]{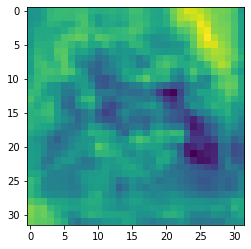}
	\includegraphics[width=0.18\textwidth]{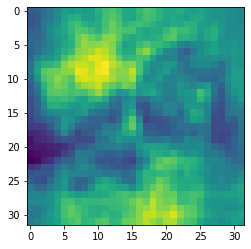} \\
	\includegraphics[width=0.18\textwidth]{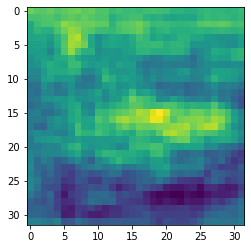}
	\includegraphics[width=0.18\textwidth]{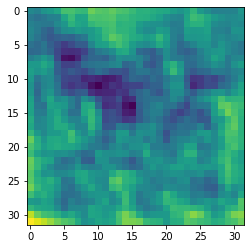}
	\includegraphics[width=0.18\textwidth]{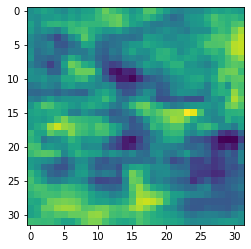}
	\includegraphics[width=0.18\textwidth]{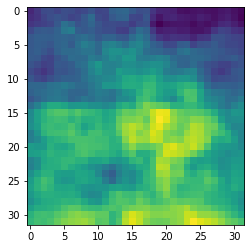}
	\includegraphics[width=0.18\textwidth]{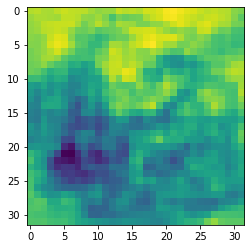} \\
	\includegraphics[width=0.18\textwidth]{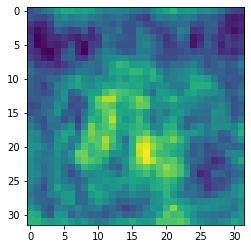}
	\includegraphics[width=0.18\textwidth]{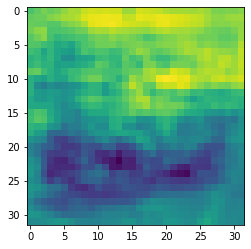}
	\includegraphics[width=0.18\textwidth]{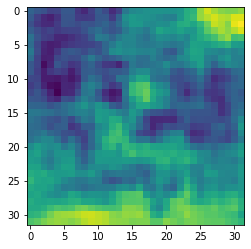}
	\includegraphics[width=0.18\textwidth]{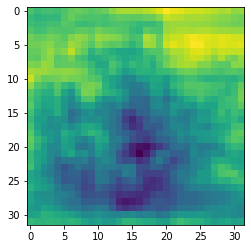}
	\includegraphics[width=0.18\textwidth]{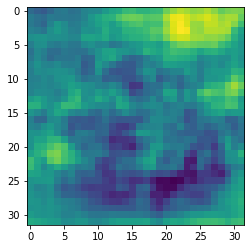} \\
	\includegraphics[width=0.18\textwidth]{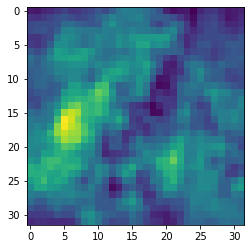}
	\includegraphics[width=0.18\textwidth]{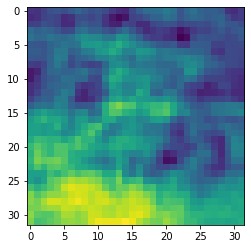}
	\includegraphics[width=0.18\textwidth]{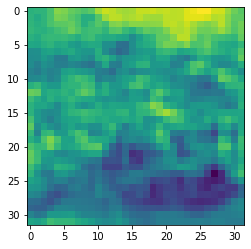}
	\includegraphics[width=0.18\textwidth]{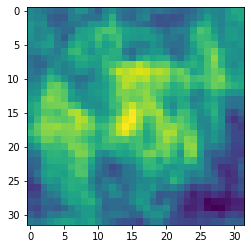}
	\includegraphics[width=0.18\textwidth]{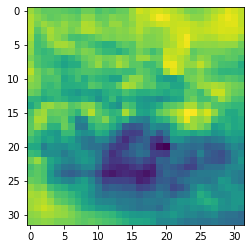}
	\caption{We show 20 augmentation data of CIFAR10 dataset here. Notice that the augmentation data is close to random noise and can not be classified as any class.}
	\label{Augmentation data in CIFAR10}
\end{figure*}

We use the data augmentation framework the same as FedMix, as shown in Algorithm \ref{Construct Augmentation Data}. For each local dataset, we upload the mean of each $M$ samples to the server. The constructed augmentation data is close to random noise. As shown in Figure \ref{Augmentation data in CIFAR10}, we randomly choose one sample in the augmentation dataset of CIFAR10 dataset.

% \begin{wrapfigure}{R}{0.5\textwidth}
% \vspace{2.em}
% \begin{minipage}{0.51\textwidth}
\begin{algorithm}[h]
	\small
	% \begin{algorithm}[!t]
	% 	\small
	\begin{algorithmic}[1]
		\Require{local Datasets $D_1, \dots, D_N$, number of augmentation data for each client $K$, number of samples to construct one augmentation sample $M$.}
		\Ensure{Augmentation Dataset $D_{p}$.}
		\myState{Initialize $D_{p} = \emptyset$.}
		\For{$i = 1, \dots, N$}
		\For{$k = 1, \dots, K$}
		\myState{Randomly sample $x_1, \dots, x_M$ from $D_i$.}
		\myState{$\bar{x} = \frac{1}{M} \sum_{m=1}^{M} x_M$.}
		\myState{$D_p = D_p \cup \{\bar{x}\}$}
		\EndFor
		\EndFor
	\end{algorithmic}
	\mycaptionof{algorithm}{\small Construct Augmentation Data}
	\label{Construct Augmentation Data}
	% \end{algorithm}
\end{algorithm}
\begin{algorithm}[h]
	\small
	% \begin{algorithm}[!t]
	% 	\small
	\begin{algorithmic}[1]
		\Require{Proxy Datasets $D_{prox}$, number of augmentation data $K$, number of samples to construct one augmentation sample $M$.}
		\Ensure{Augmentation Dataset $D_{p}$.}
		\myState{Initialize $D_{p} = \emptyset$.}
		\For{$k = 1, \dots, K$}
		\myState{Randomly sample $x_1, \dots, x_M$ from $D_{prox}$.}
		\myState{$\bar{x} = \frac{1}{M} \sum_{m=1}^{M} x_M$.}
		\myState{$D_p = D_p \cup \{\bar{x}\}$}
		\EndFor
	\end{algorithmic}
	\mycaptionof{algorithm}{\small Construct Augmentation Data by Proxy Data}
	\label{Construct Augmentation Data by Proxy Data}
	% \end{algorithm}

\end{algorithm}

% \begin{wrapfigure}{R}{0.31\textwidth}
% \begin{minipage}{0.51\textwidth}
% \end{wrapfigure}

\section{Additional Results}

\label{sec:Additional-Results}

\begin{figure*}
	\centering
	\subfigure[\small Convergence curve on RotatedMNIST]{\includegraphics[width=0.32\textwidth]{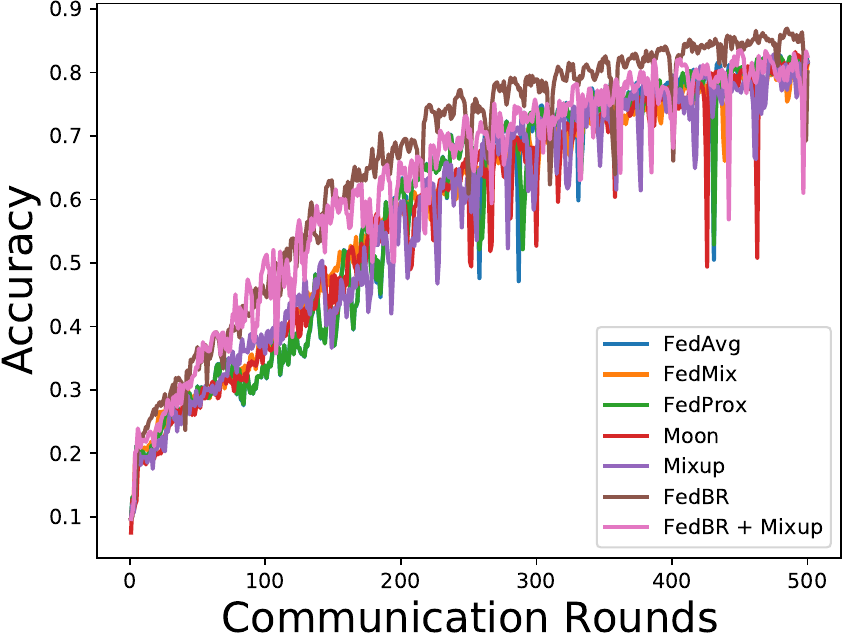}}
	\subfigure[\small Convergence curve on CIFAR10]{\includegraphics[width=0.32\textwidth]{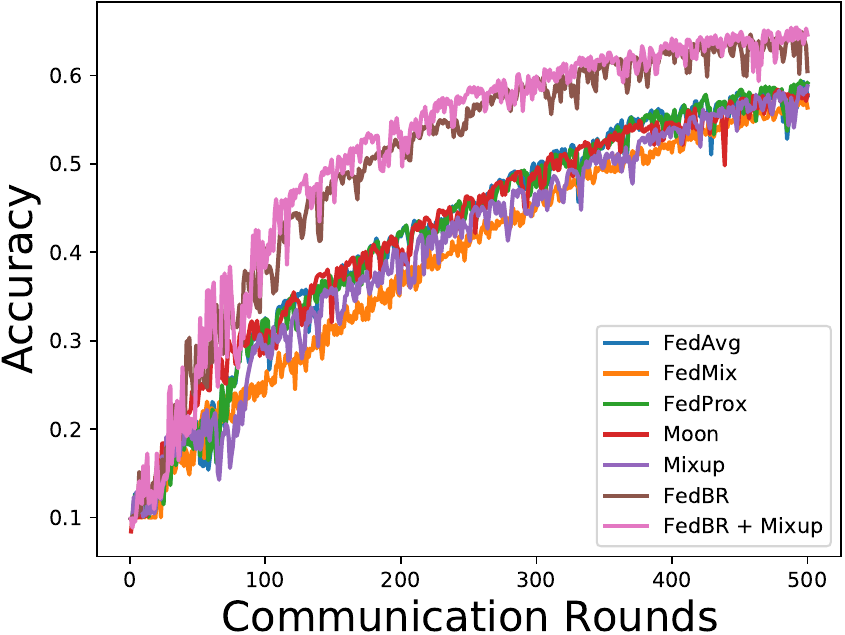}}
	\subfigure[\small Convergence curve on CIFAR100]{\includegraphics[width=0.32\textwidth]{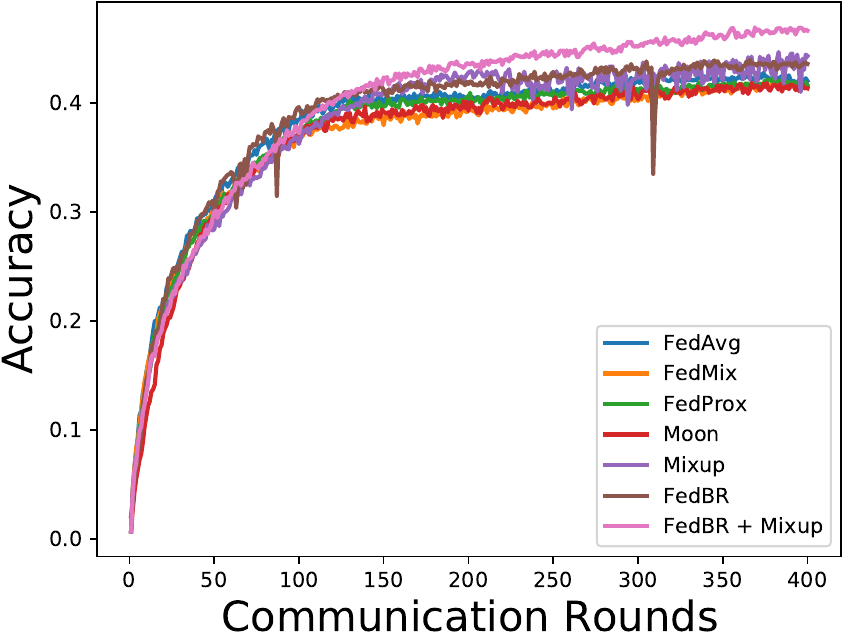}}
	\vspace{-1em}
	\caption{\small Convergence curve of algorithms on different datasets.}
	\label{convergence curve appendix}
\end{figure*}

\label{sec:Additional Results}

\subsection{Results with Error Bar}

In this section, we report the performance of our method FedAug and other baselines with an error bar to verify the performance gain of our proposed method.

\begin{table*}[!ht]
	\small
	\centering
	% 	\vspace{-1em}
	\caption{\small
		\textbf{Performance of algorithms with error bar.}
		All examined algorithms use FedAvg as the backbone. We run 1000 communication rounds on RotatedMNIST and CIFAR10 for each algorithm. For each algorithm, we run three different trials with different random seeds. For each trial, we report the mean of maximum 5 accuracies for test datasets and the number of communication rounds to reach the threshold accuracy.
		% Only very small part of combinations in Test 3 can be found in Train sets.
	}
	\vspace{-0.75em}
	\label{Performance of algorithms with error bar}
	% \resizebox{1.\textwidth}{!}{%
	\begin{tabular}{l c c c c c c c c c c}
		\toprule
		\multirow{2}{*}{Algorithm} & \multicolumn{2}{c}{RotatedMNIST} & \multicolumn{2}{c}{CIFAR10}                                                       \\
		\cmidrule(lr){2-3} \cmidrule(lr){4-5}
		                           & Acc (\%)                         & Rounds for 80\%             & Acc (\%)                      & Rounds for 55\%     \\
		\midrule
		ERM (FedAvg)               & $82.78 \pm 0.38$                 & 821 (1.0X)                  & $58.97 \pm 0.30$              & 742 (1.0X)          \\
		% 			FedProx         & 82.32 & 824 (1.0X) & 59.14 & 738 (1.0X)  \\
		% 			SCAFFOLD        & 82.49 & 814 (1.0X) & 59.00 & 738 (1.0X)  \\ 
		% 			FedMix          & 81.33 & 902 (0.9X) & $57.33 \pm 0.10$ & 883 (0.8X)  \\
		% 			Moon            & 82.68 & 864 (0.9X) & $58.23 \pm 0.12$ & 798 (0.9X)  \\
		DANN                       & $84.67 \pm 0.46$                 & 754 (1.1X)                  & $58.98 \pm 0.61$              & 747 (1.0X)          \\
		Mixup                      & $82.38 \pm 0.07$                 & 853 (1.0X)                  & $58.32 \pm 0.33$              & 822 (0.9X)          \\
		GroupDRO                   & $80.65 \pm 0.53$                 & 929 (0.9X)                  & $56.72 \pm 0.26$              & 840 (0.9X)          \\
		\midrule
		\algopt (Ours)             & $\boldsymbol{87.05 \pm 0.44}$    & \textbf{637 (1.3X)}         & $\boldsymbol{64.62 \pm 0.32}$ & \textbf{374 (2.0X)} \\
		% MNIST train   & 81.22 & & 82.51 & 84.88 & 82.57 & 80.35 & 84.58 &
		\bottomrule
	\end{tabular}%
	% }
	\vspace{-.75em}
\end{table*}

\begin{table*}[!ht]
	\small
	\centering
	% 	\vspace{-1em}
	\caption{\small
		\textbf{Performance of algorithms on CIFAR10.} We split CIFAR10 dataset to 10 clients with $\alpha = 0.1$, without additional rotation. For each algorithm, we run 1000 communication rounds on ResNet18 (with group normalization), and set local steps to 50. Note that we set momentum to 0.9 for ResNet18.
		% Only very small part of combinations in Test 3 can be found in Train sets.
	}
	\vspace{-0.75em}
	\label{Performance of cifar10 on resnet}
	% \resizebox{1.\textwidth}{!}{%
	\begin{tabular}{l c c c c c c c c c c}
		\toprule

		                    & FedAvg & FedProx & Moon  & VHL  & \algopt (ours) \\
		\midrule
		Accuracy (ResNet18) & 45.91  & 46.28   & 43.85 & 43.7 & 47.29          \\
		\bottomrule
	\end{tabular}%
	% }
	\vspace{-.75em}
\end{table*}

\subsection{Ablation Study of \algopt}

\paragraph{Values of $\tau_1$ and $\tau_2$ in Component 2.}
In this paragraph, we investigate how the value of $\tau_1$ and $\tau_2$ affect the performance of the second component of \algopt. In table  \ref{Performance of AugCA under different weights of AugCA-O and AugCA-S}, we show the results on Rotated-MNIST dataset with different weights $\tau_1$ and $\tau_2$. Results show that: 1) Setting $\tau_2 = 0$ , which only minimizes the distance of global and local features, has significant performance gain compare with ERM. However, adding $\tau_2$ can further improve the performance. 2)  The best weight on Rotated-MNIST dataset is $\tau_1 = 2.0$ and $\tau_2 = 0.5$.

\begin{table*}[!ht]
	\small
	\centering
	% 	\vspace{-1em}
	\caption{\small
		\textbf{Performance of Component 2 of \algopt under different values of $\tau_1, \tau_2$.}
		We run 1000 communication rounds on RotatedMNIST dataset. For each setting, we run three different trials with different random seeds. For each trial, we report the mean of maximum 5 accuracies for test datasets and the number of communication rounds to reach the threshold accuracy.
		% Only very small part of combinations in Test 3 can be found in Train sets.
	}
	\vspace{-0.75em}
	\label{Performance of AugCA under different weights of AugCA-O and AugCA-S}
	% \resizebox{1.\textwidth}{!}{%
	\begin{tabular}{l c c c c c c c c c c}
		\toprule

		$\tau_1$ & $\tau_2$ & Acc (\%)         & Rounds for 80\% & Rounds for 85\% \\
		\midrule
		2.0      & 0.0      & $86.11 \pm 0.77$ & 746             & 933             \\
		2.0      & 0.1      & $86.22 \pm 0.33$ & 753             & 920             \\
		2.0      & 0.5      & $87.24 \pm 0.50$ & 647             & 851             \\
		2.0      & 1.0      & $86.25 \pm 0.87$ & 705             & 922             \\
		2.0      & 2.0      & $86.01 \pm 0.33$ & 680             & 932             \\
		% MNIST train   & 81.22 & & 82.51 & 84.88 & 82.57 & 80.35 & 84.58 &
		\bottomrule
	\end{tabular}%
	% }
	\vspace{-.75em}
\end{table*}

\begin{table*}[!ht]
	\small
	\centering
	% 	\vspace{-1em}
	\caption{\small
		\textbf{Performance of \algopt under different values of $\tau_1, \tau_2$.}
		We run 1000 communication rounds on the CIFAR10 dataset. For each setting, we run three different trials with different random seeds. For each trial, we report the mean of maximum 5 accuracies for test datasets and the number of communication rounds to reach the threshold accuracy.
		% Only very small part of combinations in Test 3 can be found in Train sets.
	}
	\vspace{-0.75em}
	\label{Performance of FedAug under different weights of AugCA-O and AugCA-S}
	% \resizebox{1.\textwidth}{!}{%
	\begin{tabular}{l c c c c c c c c c c}
		\toprule

		$\tau_1$ & $\tau_2$ & Acc (\%)         & Rounds for 55\% & Rounds for 60\% \\
		\midrule
		2.0      & 0.0      & $64.05 \pm 0.27$ & 390             & 563             \\
		2.0      & 0.5      & $64.26 \pm 0.47$ & 382             & 585             \\
		2.0      & 1.0      & $64.77 \pm 0.24$ & 374             & 533             \\
		2.0      & 2.0      & $64.62 \pm 0.32$ & 374             & 541             \\
		% MNIST train   & 81.22 & & 82.51 & 84.88 & 82.57 & 80.35 & 84.58 &
		\bottomrule
	\end{tabular}%
	% }
	\vspace{-.75em}
\end{table*}

\paragraph{Weights of the first component of \algopt.} In this paragraph, we investigate how the weights of the first component of \algopt affect the performance of models in table \ref{Performance of AugMean under different weights}.

\begin{table*}[!ht]
	\small
	\centering
	% 	\vspace{-1em}
	\caption{\small
		\textbf{Performance of component 1 under different weights.}
		We run 1000 communication rounds on the CIFAR10 dataset. For each setting, we run three different trials with different random seeds. For each trial, we report the mean of maximum 5 accuracies for test datasets and the number of communication rounds to reach the threshold accuracy. We use $\lambda$ as the weight of the first component of \algopt.
		% Only very small part of combinations in Test 3 can be found in Train sets.
	}
	\vspace{-0.75em}
	\label{Performance of AugMean under different weights}
	% \resizebox{1.\textwidth}{!}{%
	\begin{tabular}{l c c c c c c c c c c}
		\toprule

		$\lambda$ & Acc (\%)         & Rounds for 55\% & Rounds for 60\% \\
		\midrule
		0.1       & $64.12 \pm 0.27$ & 442             & 591             \\
		0.5       & $64.92 \pm 0.46$ & 385             & 536             \\
		1.0       & $64.50 \pm 0.34$ & 379             & 565             \\
		% MNIST train   & 81.22 & & 82.51 & 84.88 & 82.57 & 80.35 & 84.58 &
		\bottomrule
	\end{tabular}%
	% }
	\vspace{-.75em}
\end{table*}

\paragraph{Domain robustness of FL and DG algorithms.}
We also hope that our method can increase the model's robustness because it expects to train client invariant features.
Therefore, we calculate the worst accuracy on test datasets of all clients/domains and report the mean of each algorithm's top 5 worst accuracies in Table \ref{Worst Case Performance of algorithms} to show the domain robustness of algorithms.
We have the following findings:
1) \algopt significantly outperforms other approaches, and the improvements of \algopt over FedAvg become more significant than the mean accuracy in Table \ref{Performance of algorithms}. \algopt has a role in learning a domain-invariant feature and improving robustness, as evidenced by this finding.
2) Under these settings, DG baselines outperform FedAvg. This finding demonstrates that the DG algorithms help to enhance domain robustness.

\begin{table*}[!ht]
	\small
	\centering
	% 	\vspace{-1em}
	\caption{\small
		\textbf{Performance of algorithms.}
		All examined algorithms use FedAvg as the backbone. We run 1000 communication rounds on RotatedMNIST and CIFAR10 for each algorithm, 800 communication rounds CIFAR100 and 400 communication rounds for PACS. We report the mean of maximum 5 accuracies for test datasets and the number of communication rounds to reach the final accuracy of ERM .
		% Only very small part of combinations in Test 3 can be found in Train sets.
	}
	\vspace{-0.75em}
	\label{Performance of algorithms-appendix}
	% \resizebox{1.\textwidth}{!}{%
	\begin{tabular}{l c c c c c c c c c c c c}
		\toprule
		\multirow{2}{*}{Algorithm} & \multicolumn{2}{c}{RotatedMNIST} & \multicolumn{2}{c}{CIFAR10} & \multicolumn{2}{c}{PACS}                                                              \\
		\cmidrule(lr){2-3} \cmidrule(lr){4-5} \cmidrule(lr){6-7}
		                           & Acc (\%)                         & Rounds (Speed up)           & Acc (\%)                 & Rounds (Speed up)   & Acc (\%)       & Rounds (Speed up)   \\
		\midrule
		ERM (FedAvg)               & 82.47                            & 828 (1.0X)                  & 58.99                    & 736 (1.0X)          & 64.03          & 168 (1.0X)          \\
		FedProx                    & 82.32                            & 824 (1.0X)                  & 59.14                    & 738 (1.0X)          & 65.10          & 168 (1.0X)          \\
		SCAFFOLD                   & 82.49                            & 814 (1.0X)                  & 59.00                    & 738 (1.0X)          & 64.49          & 168 (1.0X)          \\
		FedMix                     & 81.33                            & 902 (0.9X)                  & 57.37                    & 872 (0.8X)          & 62.14          & 228 (0.7X)          \\
		Moon                       & 82.68                            & 864 (0.9X)                  & 58.23                    & 820 (0.9X)          & 64.86          & 122 (1.4X)          \\
		DANN                       & 84.83                            & 743 (1.1X)                  & 58.29                    & 782 (0.9X)          & 64.97          & 109 (1.5X)          \\
		Mixup                      & 82.56                            & 840 (1.0X)                  & 58.57                    & 826 (0.9X)          & 64.36          & 210 (0.8X)          \\
		GroupDRO                   & 80.23                            & 910 (0.9X)                  & 56.57                    & 835 (0.9X)          & 64.40          & 170 (1.0X)          \\
		\midrule
		\algopt (Ours)              & \textbf{86.58}                   & \textbf{628 (1.3X)}         & \textbf{64.65}           & \textbf{496 (1.5X)} & \textbf{65.63} & \textbf{100 (1.7X)} \\
		% MNIST train   & 81.22 & & 82.51 & 84.88 & 82.57 & 80.35 & 84.58 &
		\bottomrule
	\end{tabular}%
	% }
	\vspace{-.75em}
\end{table*}

\begin{table*}[!ht]
	\small
	\centering
	% 	\vspace{-1em}
	\caption{\small
		\textbf{Worst Case Performance of algorithms.}
		All examined algorithms use FedAvg as the backbone. We run 1000 communication rounds on RotatedMNIST and CIFAR10 for each algorithm, 800 rounds for CIFAR100, and 400 communication rounds for PACS. We calculate the worst accuracy for all clients in each round and report the mean of the top 5 worst accuracies for each method. Besides, we report the number of communication rounds to reach the final worst accuracy of FedAvg.
		% Only very small part of combinations in Test 3 can be found in Train sets.
	}
	\vspace{-0.75em}
	\label{Worst Case Performance of algorithms}
	% \resizebox{1.\textwidth}{!}{%
	\begin{tabular}{l c c c c c c c c c c c c}
		\toprule
		\multirow{2}{*}{Algorithm} & \multicolumn{2}{c}{RotatedMNIST} & \multicolumn{2}{c}{CIFAR10} & \multicolumn{2}{c}{PACS}                                                             \\
		\cmidrule(lr){2-3} \cmidrule(lr){4-5} \cmidrule(lr){6-7}
		                           & Acc (\%)                         & Rounds (Speed up)           & Acc (\%)                 & Rounds (Speed up)   & Acc (\%)       & Rounds (Speed up)  \\
		\midrule
		ERM (FedAvg)               & 66.60                            & 816 (1.0X)                  & 41.30                    & 846 (1.0X)          & 42.79          & 170 (1.0X)         \\
		FedProx                    & 65.88                            & 780 (1.0X)                  & 41.84                    & 840 (1.0X)          & 42.82          & 170 (1.0X)         \\
		SCAFFOLD                   & 66.72                            & 804 (1.0X)                  & 40.88                    & 840 (1.0X)          & 41.63          & 170 (1.0X)         \\
		FedMix                     & 60.52                            & 910 (0.9X)                  & 28.44                    & -                   & 38.00          & -                  \\
		Moon                       & 66.18                            & 866 (0.9X)                  & 40.34                    & 908 (0.9X)          & 41.59          & 66 (2.6X)          \\
		DANN                       & 67.85                            & 753 (1.1X)                  & 43.38                    & 747 (1.1X)          & 40.51          & 59 (2.9X)          \\
		Mixup                      & 66.25                            & 836 (1.0X)                  & 40.32                    & 984 (0.9X)          & 41.89          & 252 (0.7X)         \\
		GroupDRO                   & 68.53                            & \textbf{568 (1.4X)}         & 46.90                    & 656 (1.3X)          & 43.18          & 246 (0.7X)         \\
		\midrule
		% 			Ours \\
		\algopt (Ours)             & \textbf{77.13}                   & 630 (1.3X)                  & \textbf{48.94}           & \textbf{632 (1.3X)} & \textbf{43.99} & \textbf{58 (2.9X)} \\
		% 			AugMean & 68.55 & 738 & 48.26 & \textbf{572} & 43.00 & 52\\
		% 			AugCA & 76.88 & 614 & 45.42 & 798 & \\
		% MNIST train   & 81.22 & & 82.51 & 84.88 & 82.57 & 80.35 & 84.58 &
		\bottomrule
	\end{tabular}%
	% }
	\vspace{-.75em}
\end{table*}

\subsection{T-SNE and Classcifier Outputs of Toy Examples}

\label{sec:t-SNE and classcifier output}

\begin{figure*}
	\centering
	\subfigure[\small Global feature of $X_1$, $F_g(X_1)$]{\includegraphics[width=0.32\textwidth]{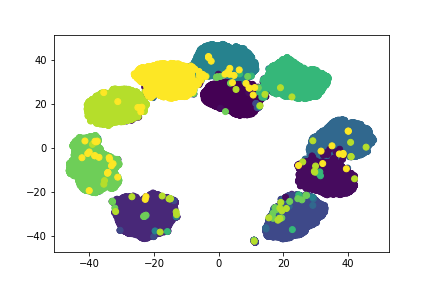}}
	\subfigure[\small Local feature of $X_1$, $F_1(X_1)$]{\includegraphics[width=0.32\textwidth]{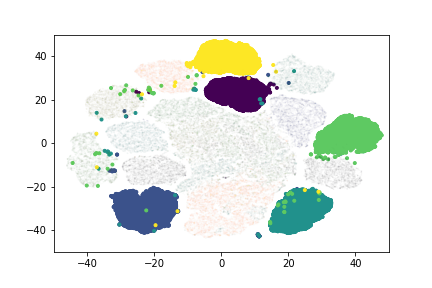}}
	\subfigure[\small Local feature of $X_2$, $F_1(X_2)$]{\includegraphics[width=0.32\textwidth]{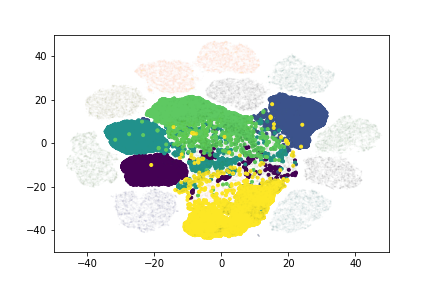}}
	\vspace{-1em}
	\caption{\small Features after the second component of \algopt.}
	\label{Features after AugCA}
\end{figure*}

\begin{figure*}
	\centering
	\includegraphics[width=0.32\textwidth]{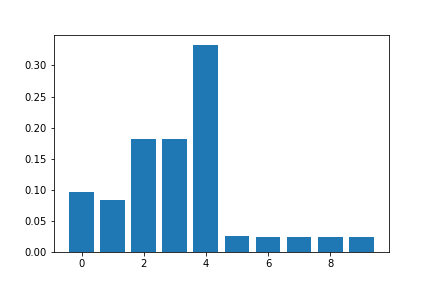}
	\vspace{-1em}
	\caption{\small Classifier output after the first component of \algopt on unseen classes.}
	\label{Classifier after AugMean}
\end{figure*}

As the setting in Figure \ref{Output of model trained on class 1-5} and Figure \ref{Class imbalance figure}, we investigate if the two components of \algopt will help for mitigating the proposed bias on feature and classifier. Figure \ref{Features after AugCA} show the features after the second component of \algopt, which implies this component can significantly mitigate the proposed feature bias: 1) on the seen datasets, local features are close to global features. 2) on the unseen datasets, the local feature is far away from that of seen datasets.
Figure \ref{Classifier after AugMean} shows the output of the local classifier after the first component of \algopt on unseen classes. Notice that compared with Figure \ref{Class imbalance figure}, the output is more balanced.

\begin{figure*}
	\centering
	\subfigure[\small Global feature of $X_1$, $F_g(X_1)$]{\includegraphics[width=0.32\textwidth]{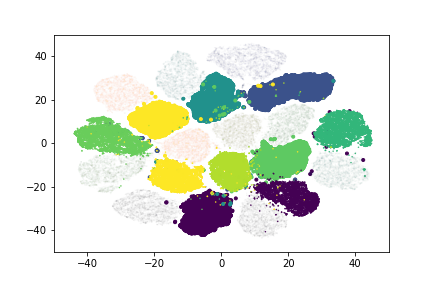}}
	\subfigure[\small Fine-tuned Local feature of $X_1$, $F_1(X_1)$]{\includegraphics[width=0.32\textwidth]{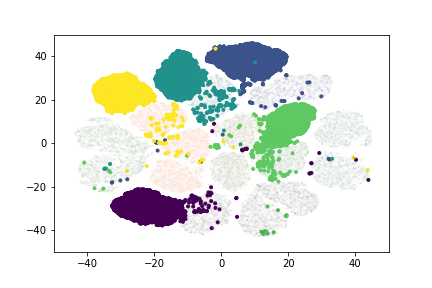}}
	\subfigure[\small Fine-tuned Local feature of $X_2$, $F_1(X_2)$]{\includegraphics[width=0.32\textwidth]{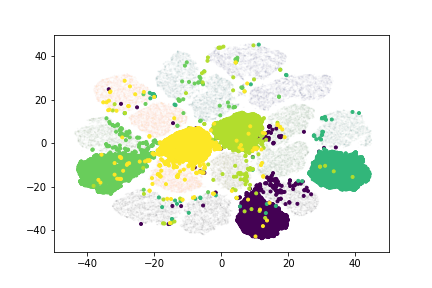}}
	\vspace{-1em}
	\caption{\small Fine-tuned local features after 10 local epochs.}
	\label{Fine-tuned Features after 10 local epochs}
\end{figure*}

\begin{figure*}
	\centering
	\subfigure[\small Global feature of $X_1$, $F_g(X_1)$]{\includegraphics[width=0.32\textwidth]{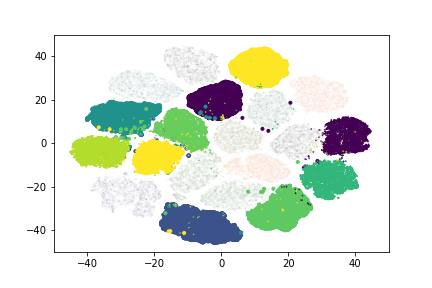}}
	\subfigure[\small Fine-tuned Local feature of $X_1$, $F_1(X_1)$]{\includegraphics[width=0.32\textwidth]{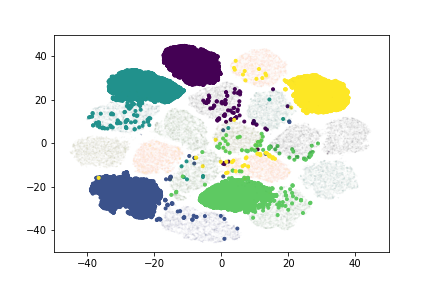}}
	\subfigure[\small Fine-tuned Local feature of $X_2$, $F_1(X_2)$]{\includegraphics[width=0.32\textwidth]{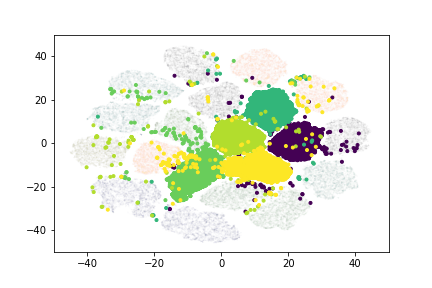}}
	\vspace{-1em}
	\caption{\small Fine-tuned local features after 20 local epochs.}
	\label{Fine-tuned Features after 20 local epochs}
\end{figure*}

In Figure~\ref{Fine-tuned Features after 10 local epochs} and Figure~\ref{Fine-tuned Features after 20 local epochs}, we show the local learning bias when local model has better feature initialization. We copy the feature extractor of global model to local models, and randomly initialize local classifiers. Results show that: 1) The drifts between global and local features are still significant even has a good feature initialization.
2) The local features of unseen data are less relevant to the local features of seen data compare with training from scratch. This indicates that such a problem will be mitigated after enough training rounds.
3) The drifts between global and local features increase as the number of local epochs increases.

\begin{figure*}
	\centering
	\subfigure[\small Global feature of $X_1$, $F_g(X_1)$]{\includegraphics[width=0.32\textwidth]{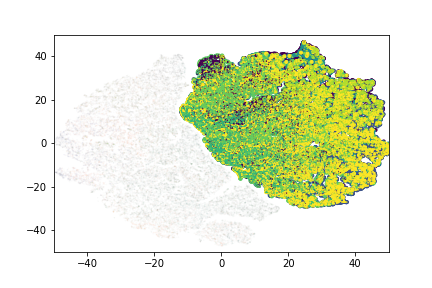}}
	\subfigure[\small Fine-tuned Local feature of $X_1$, $F_1(X_1)$]{\includegraphics[width=0.32\textwidth]{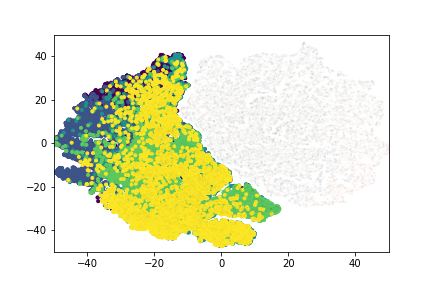}}
	\subfigure[\small Fine-tuned Local feature of $X_2$, $F_1(X_2)$]{\includegraphics[width=0.32\textwidth]{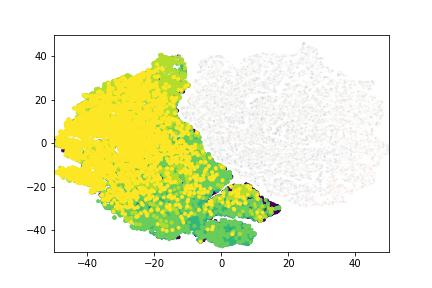}} \\
     \subfigure[\small Global feature of $X_1$, $F_g(X_1)$]{\includegraphics[width=0.32\textwidth]{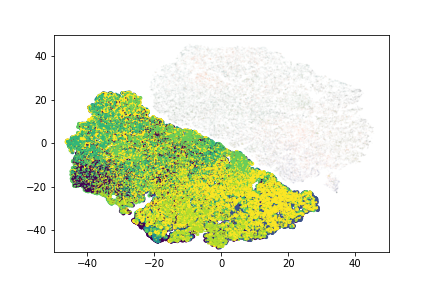}}
    	\subfigure[\small Fine-tuned Local feature of $X_1$, $F_1(X_1)$]{\includegraphics[width=0.32\textwidth]{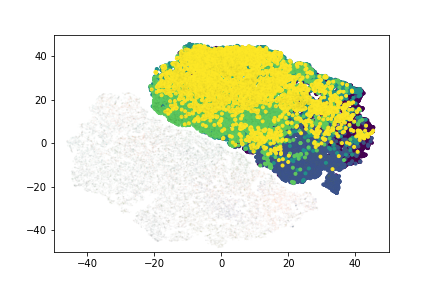}}
    	\subfigure[\small Fine-tuned Local feature of $X_2$, $F_1(X_2)$]{\includegraphics[width=0.32\textwidth]{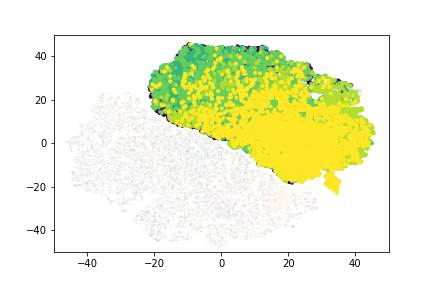}} \\
     \subfigure[\small Global feature of $X_1$, $F_g(X_1)$]{\includegraphics[width=0.32\textwidth]{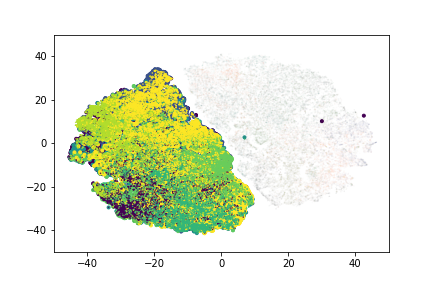}}
	\subfigure[\small Fine-tuned Local feature of $X_1$, $F_1(X_1)$]{\includegraphics[width=0.32\textwidth]{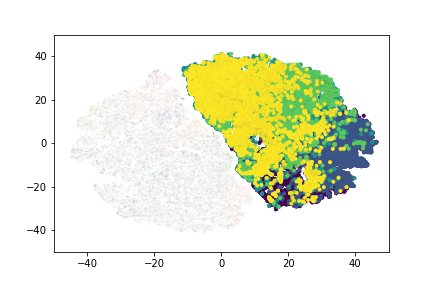}}
	\subfigure[\small Fine-tuned Local feature of $X_2$, $F_1(X_2)$]{\includegraphics[width=0.32\textwidth]{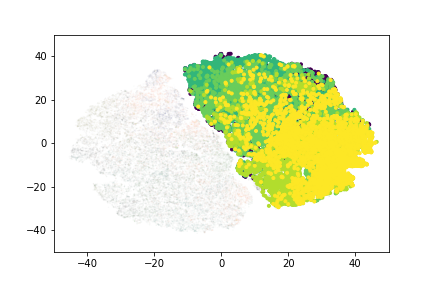}}
	\vspace{-1em}
	\caption{\small First train global model on the whole dataset for 1, 3, and 10 epoch (w.r.t. each row), then report local features after 10 local epochs.}
	\label{Fine-tuned E1 Features after 10 local epochs}
\end{figure*}

We also investigate if our observation remains for different stages of global models. In this experiment, we use CIFAR10 dataset, and train global model for 1, 3, 10 epochs on the whole dataset to obtain 29.74\%, 38.65\%, 49.28\% global accuracy, then we directly copy global models to clients (including classifier). We fine-tune the global models for 10 local epochs, results are shown in Figure~\ref{Fine-tuned E1 Features after 10 local epochs}. Results show that: For not well-trained global models, difference between global features on the same input and similarity between local features of different inputs are both significant.

% \begin{figure*}
% 	\centering
% 	\subfigure[\small Global feature of $X_1$, $F_g(X_1)$]{\includegraphics[width=0.32\textwidth]{figures/global-E3.png}}
% 	\subfigure[\small Fine-tuned Local feature of $X_1$, $F_1(X_1)$]{\includegraphics[width=0.32\textwidth]{figures/local-1-E3.png}}
% 	\subfigure[\small Fine-tuned Local feature of $X_2$, $F_1(X_2)$]{\includegraphics[width=0.32\textwidth]{figures/local-2-E3.png}}
% 	\vspace{-1em}
% 	\caption{\small First train global model on the whole dataset for 3 epochs, then report local features after 10 local epochs.}
% 	\label{Fine-tuned E3 Features after 10 local epochs}
% \end{figure*}

% \begin{figure*}
% 	\centering
% 	\subfigure[\small Global feature of $X_1$, $F_g(X_1)$]{\includegraphics[width=0.32\textwidth]{figures/global-E10.png}}
% 	\subfigure[\small Fine-tuned Local feature of $X_1$, $F_1(X_1)$]{\includegraphics[width=0.32\textwidth]{figures/local-1-E10.png}}
% 	\subfigure[\small Fine-tuned Local feature of $X_2$, $F_1(X_2)$]{\includegraphics[width=0.32\textwidth]{figures/local-2-E10.png}}
% 	\vspace{-1em}
% 	\caption{\small First train global model on the whole dataset for 10 epochs, then report local features after 10 local epochs.}
% 	\label{Fine-tuned E10 Features after 10 local epochs}
% \end{figure*}

\begin{figure}
    \centering
    \subfigure[Global, $\alpha = 0.05$]{
    \includegraphics[width=0.23\textwidth]{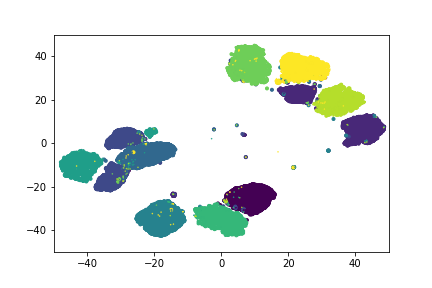}
    }
    \subfigure[Local $X_1$, $\alpha = 0.05$]{
    \includegraphics[width=0.23\textwidth]{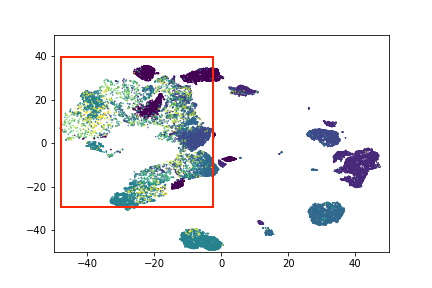}
    }
    \subfigure[Local $X_2$, $\alpha = 0.05$]{
    \includegraphics[width=0.23\textwidth]{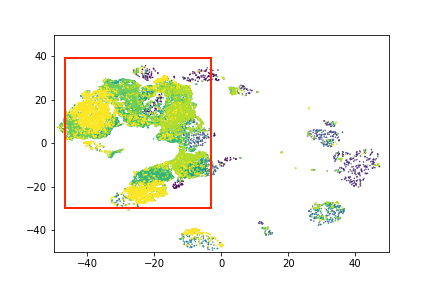}
    }
    \subfigure[Aggregate, $\alpha = 0.05$]{
    \includegraphics[width=0.23\textwidth]{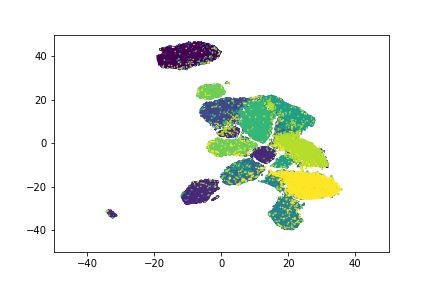}
    } \\
    \subfigure[Global, $\alpha = 0.1$]{
    \includegraphics[width=0.23\textwidth]{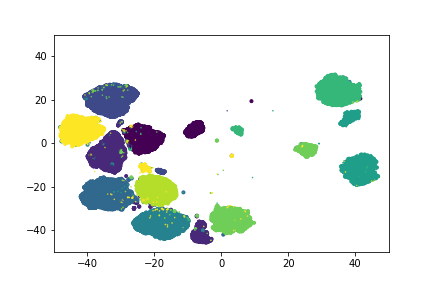}
    }
    \subfigure[Local $X_1$, $\alpha = 0.1$]{
    \includegraphics[width=0.23\textwidth]{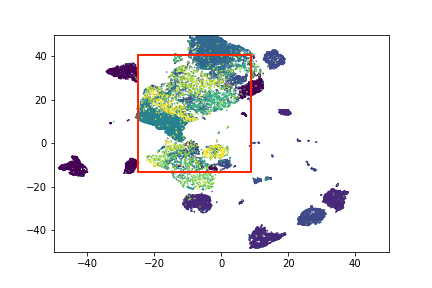}
    }
    \subfigure[Local $X_2$, $\alpha = 0.1$]{
    \includegraphics[width=0.23\textwidth]{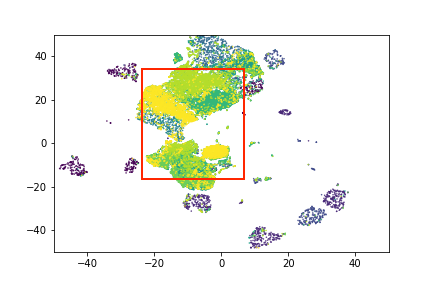}
    }
    \subfigure[Aggregate, $\alpha = 0.1$]{
    \includegraphics[width=0.23\textwidth]{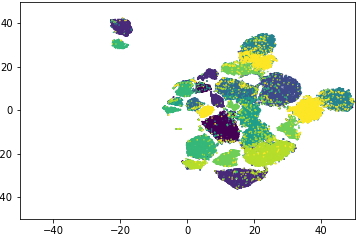}
    } \\
    \subfigure[Global, $\alpha = 0.4$]{
    \includegraphics[width=0.23\textwidth]{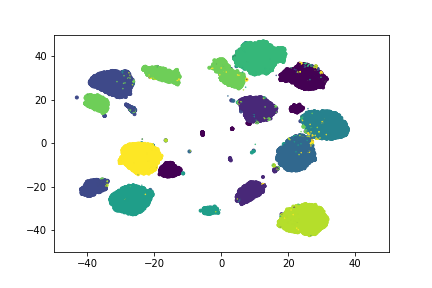}
    }
    \subfigure[Local $X_1$, $\alpha = 0.4$]{
    \includegraphics[width=0.23\textwidth]{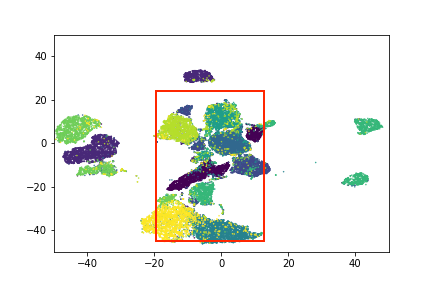}
    }
    \subfigure[Local $X_2$, $\alpha = 0.4$]{
    \includegraphics[width=0.23\textwidth]{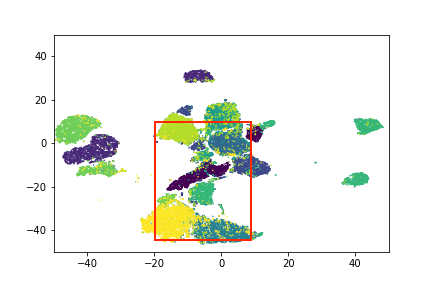}
    }
    \subfigure[Aggregate, $\alpha = 0.4$]{
    \includegraphics[width=0.23\textwidth]{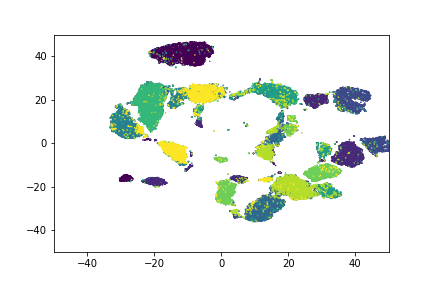}
    }
    \subfigure[Local Classifier, $\alpha = 0.05$]{
    \includegraphics[width=0.31\textwidth]{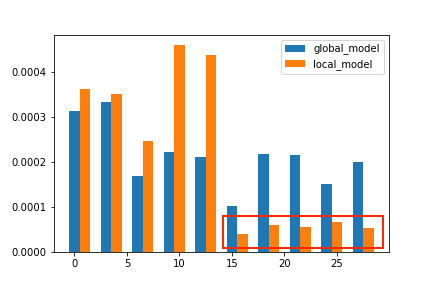}
    }
    \subfigure[Local Classifier, $\alpha = 0.1$]{
    \includegraphics[width=0.31\textwidth]{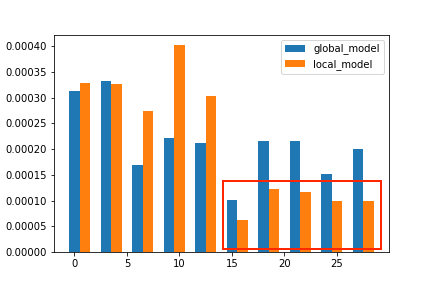}
    }
    \subfigure[Local Classifier, $\alpha = 0.4$]{
    \includegraphics[width=0.31\textwidth]{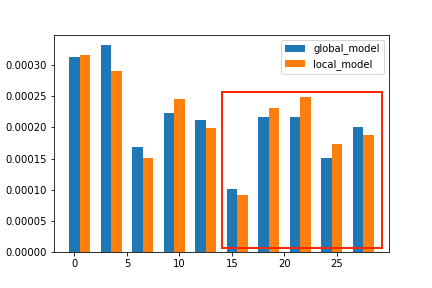}
    }
    \caption{\textbf{Illustration of our observation under mild split conditions:} We introduce a parameter $\alpha \in [0, 0.5]$ to control the level of non-i.i.d. of clients, where a larger $\alpha$ indicates less non-i.i.d., and $\alpha = 0.5$ indicates a balanced local distribution. We present the global feature, local feature on seen ($X_1$) and unseen ($X_2$) data, as well as the feature of the aggregated model. Additionally, we illustrate the output distribution of the global and local classifiers on the test data with a balanced label distribution.}
    \label{fig:Illustration of our observation under mild split conditions}
\end{figure}

\subsection{T-SNE Results on Mild Conditions}

 We introduce a parameter $\alpha \in [0, 0.5]$ to control the level of non-i.i.d. of clients, where a larger $\alpha$ indicates less non-i.i.d., and $\alpha = 0.5$ indicates a balanced local distribution. Results are shown in Figure~\ref{fig:Illustration of our observation under mild split conditions}: 1) The local feature on the unseen data (Local $X_2$) still lacks a clear decision boundary, and the local features are close even for data from different classes. 2) The decision boundary of the aggregated model becomes clearer as $\alpha$ increases, supporting the necessity of reducing the local bias. 3) Our observation on the biased classifier still holds, where a smaller $\alpha$ leads to a more biased classifier output.

\begin{figure}
    \centering
    \subfigure[CIFAR10, VGG11]{
    \includegraphics[width=0.45\textwidth]{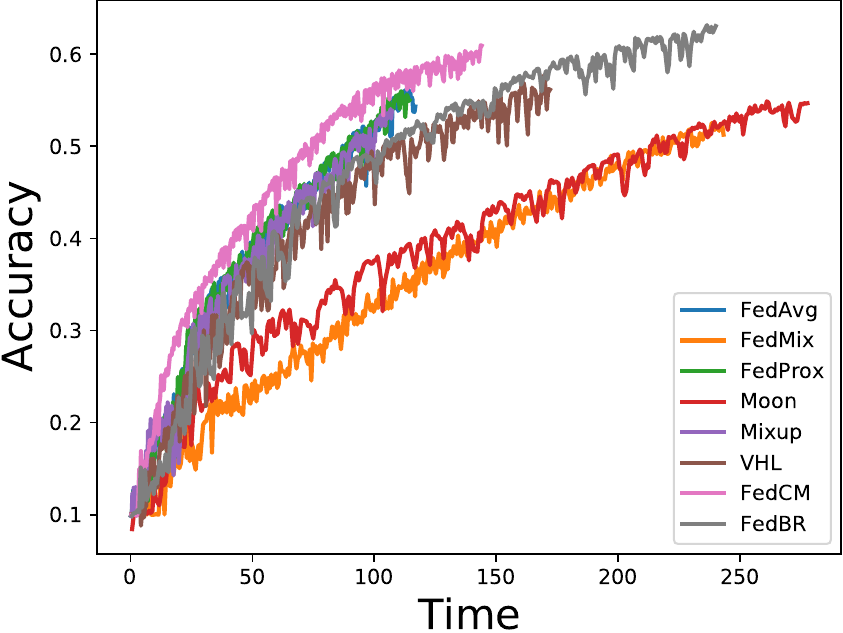}
    }
    \subfigure[CIFAR100, CCT]{
    \includegraphics[width=0.4\textwidth]{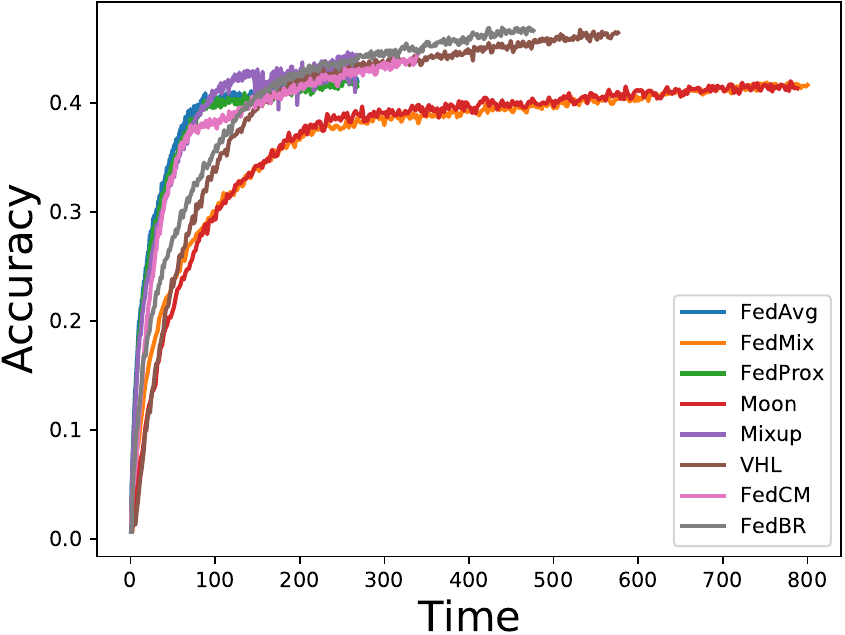}
    }
    \caption{\textbf{Convergence curve w.r.t simulation time.} We present the convergence curves of various algorithms with respect to simulation time. Our results indicate that FedBR introduces additional computational burden compared to FedAvg and FedProx. However, the computational efficiency of FedBR is comparable to that of other regularization-based baselines. }
    \label{fig:Convergence curve w.r.t simulation time}
\end{figure}

% \subsection{Illustration of min-max problem}

% In Figure~\ref{Illustration of the min-max process}, we illustrate the intuition to use the proposed min-max process. The projection layer is used to distinguish biased and unbiased features that can not be distinguished well on the original feature space, and the min step is to learn unbiased local features that close to unbiased features on the projected spaces.

% \begin{figure*}
% 	\centering
% 	{\includegraphics[width=1.0\textwidth]{./figures/min-max.jpg}}
% 	\caption{\small Illustration of the min-max process. In the left side of the figure, we show that the projection layer maximizes the distance between biased and unbiased features. In the right side of the figure, we show that the unbiased local feature is trained by forcing projected features closer to unbiased features.}
% 	\label{Illustration of the min-max process}
% \end{figure*}
%%%%%%%%%%%%%%%%%%%%%%%%%%%%%%%%%%%%%%%%%%%%%%%%%%%%%%%%%%%%
\end{document}